\documentclass{article}

\usepackage[preprint]{neurips_2025}

\usepackage[utf8]{inputenc} %
\usepackage[T1]{fontenc}    %
\usepackage{hyperref}       %
\usepackage{url}            %
\usepackage{booktabs}       %
\usepackage{amsfonts}       %
\usepackage{nicefrac}       %
\usepackage{microtype}      
\usepackage{algorithm}
\usepackage{algpseudocode}

\floatname{algorithm}{Algorithm}

\usepackage{graphicx}
\usepackage{amsmath}
\usepackage{amssymb}
\usepackage{booktabs}
\usepackage{float}
\usepackage{booktabs}
\usepackage{color}
\usepackage[dvipsnames,table,xcdraw]{xcolor}

\newcommand\comment[1]{$\textcolor{red}{\bullet}$ }
\def\method{SAIL}

\definecolor{sdf}{RGB}{230, 255, 230}
\definecolor{mvs}{RGB}{230, 240, 255}

\definecolor{best}{RGB}{255, 200, 200}
\definecolor{second}{RGB}{255, 220, 180} %
\definecolor{third}{RGB}{255, 250, 200} %

\usepackage{overpic}
\usepackage{multirow}
\usepackage{stfloats}
\usepackage{soul}
\usepackage{wrapfig}
\usepackage{graphicx}
\usepackage{pict2e}
\usepackage{colortbl}
\usepackage{amssymb}

\usepackage[T1]{fontenc}    
\usepackage{newunicodechar}
\newunicodechar{↔}{\ensuremath{\leftrightarrow}}
\usepackage{pifont}

\title{SAIL: Self-supervised Albedo Estimation from Real Images with a Latent Diffusion Model}

\author{
    Hala Djeghim$^
    {1,2}$ \qquad
    Nathan Piasco$^{1}$ \qquad
    Luis Roldão$^{1}$ \qquad
    Moussab Bennehar$^{1}$ \qquad \\
    \textbf{Dzmitry Tsishkou}$^{1}$ \qquad
    \textbf{Céline Loscos}$^{3}$  \qquad 
    \textbf{Désiré Sidibé}$^{2}$ 
    \\
    $^{1}$Noah's Ark, Huawei Paris Research Center, France \\
    $^{2}$ IBISC, Universit\'e Paris-Saclay, Univ Evry, France\\
    $^{3}$ L Research, France\\
    }

\begin{document}

\maketitle

\begin{figure*}[h!] %
  \centering
  \includegraphics[width=0.9\textwidth]{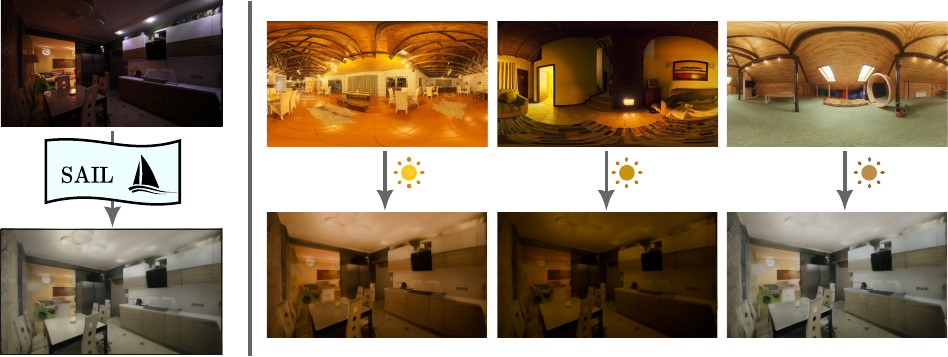} %
  \caption{\textbf{Self-Supervised Albedo Estimation from Real Images -- } From a single image under real-world lighting conditions, \method~ extracts high-fidelity albedo by repurposing and finetuning a pretrained latent diffusion model (left). The estimated albedo enables downstream tasks such as single-image virtual relighting, demonstrated using Blender~\cite{blender} with different environment maps (right).}
  \label{fig:teaser}
\end{figure*}

\begin{abstract}
Intrinsic image decomposition aims at separating an image into its underlying albedo and shading components, isolating the base color from lighting effects to enable downstream applications such as virtual relighting and scene editing. 
Despite the rise and success of learning-based approaches, intrinsic image decomposition from real-world images remains a significant challenging task due to the scarcity of labeled ground-truth data.
Most existing solutions rely on synthetic data as supervised setups, limiting their ability to generalize to real-world scenes. Self-supervised methods, on the other hand, often produce albedo maps that contain reflections and lack consistency under different lighting conditions.
To address this, we propose \method, an approach designed to estimate albedo-like representations from single-view real-world images. We repurpose the prior knowledge of a latent diffusion model for unconditioned scene relighting as a surrogate objective for albedo estimation. To extract the albedo, we introduce a novel intrinsic image decomposition fully formulated in the latent space.
To guide the training of our latent diffusion model, we introduce regularization terms that constrain both the lighting-dependent and independent components of our latent image decomposition.
\method~predicts stable albedo under varying lighting conditions and generalizes to multiple scenes, using only unlabeled multi-illumination data available online.
\end{abstract}
    
\section{Introduction}

Decomposing an image into its inherent lighting-invariant properties and those dependent on lighting has been a significant long-standing challenge for decades.
This decomposition is essential for various applications such as scene editing, relighting, and object insertion.
Although significant progress has been made with synthetic datasets, the problem remains difficult for real-world images.
In natural environments, objects and scene elements are visible only because of their interaction with light, which introduces ambiguity in the decomposition process. As a result, estimating albedo maps without illumination effects from in-the-wild images remains an ill-posed problem.

Recent data-driven solutions have shown impressive results in disentangling geometry, material properties, and lighting conditions~\cite{zhu2022learning, li2020inverse}, but often fail to capture high-frequency details as they tend to converge to an average solution.
Diffusion-based methods~\cite{kocsis2023intrinsic, Zeng_2024}, by leveraging strong prior knowledge and modeling a probabilistic distribution, better capture the diversity of valid decompositions. However, existing diffusion-based solutions rely on synthetic datasets with ground-truth supervision, limiting their ability to generalize to unseen real-world scenes, often resulting in unrealistic albedo maps that alter object properties.

Early methods such as~\cite{li2018learning} leverage time-lapse sequences to learn intrinsic decomposition in a self-supervised manner directly in image space, but struggle to separate lighting effects from the base color, resulting in poor albedo quality under strong lighting changes. More recently, Zhang et al.~\cite{zhang2024latent} proposed a latent-space approach by training a relighting model, where an albedo representation emerges implicitly. While this improves results compared to image-space methods, it still struggles in complex real-world scenes and fails to fully remove reflections on glossy surfaces, due to the absence of an explicit decomposition.

To overcome these limitations, we present \method, a self-supervised approach for estimating albedo-like representations from single-view real-world images. Specifically, we leverage the prior knowledge of a latent diffusion model by repurposing it for unconditioned scene relighting. This surrogate training objective enables us to estimate albedos without any labeled data. To do so, we take inspiration from recent work on unsupervised decomposition with diffusion model~\cite{su2024compositionalimagedecompositiondiffusion} and train \method~to jointly predict albedo and lightning de-correlated components. This image intrinsic decomposition is fully represented in the latent space to leverage the expressiveness of the latent diffusion model and scale up the quality of the generated albedo images. 
We introduce regularization of the latent space for consistent albedo estimation, robust to real-world ambiguous lighting conditions. We show that \method~outperforms both supervised and self-supervised baselines by producing high-quality albedo predictions consistent and robust to real images captured under various conditions. Furthermore, we demonstrate that our representation generalizes well to out-of-domain data and outdoor scenes, and that it can be trained efficiently using available time-lapse images from the Internet. In summary, our contributions are: 

-- A latent diffusion model for predicting albedo-like representations from single-view real-world images.

-- A supervised training pipeline of a latent diffusion model for unconditioned scene relighting, as a surrogate task for self-supervised albedo-like image estimation.

-- A novel image decomposition fully represented in the latent space that explicitly separates lighting-invariant and lighting-dependent components.

-- A set of carefully designed regularization terms applied in the latent space to guide the training toward a consistent albedo estimate under various lighting conditions.

-- A comprehensive evaluation on three public datasets showing that our method outperforms supervised and unsupervised baselines in robustness and albedo quality.

\section{Related work}

\paragraph{Image intrinsic decompositions.}
Image intrinsic decomposition is a classic problem in computer vision and computer graphics. The scene intrinsic model as described in 1978~\cite{barrow1978recovering}, is composed of three components: surface reflectance, surface orientation, and incident light. Most methods focus on decomposing an image $I$ into a reflectance (albedo) component, invariant to illumination effects, and a shading component, which captures illumination effects.\\ 
Early solutions~\cite{garces2022survey} were inspired by Retinex Theory~\cite{land1971lightness}, where large intensity gradients in an image are attributed to changes in reflectance, while small gradients are due to variations in illumination.

Recent intrinsic decomposition learning-based solutions are usually part of a full inverse rendering pipeline that aims at disentangling scene geometry from its base color, lighting conditions, and material properties \cite{zhu2023i2,choi2023mair, li2020inverse}. Many learning-based methods rely on synthetic datasets because they give access to ground truth data~\cite{liu2024review}. However, models trained purely on synthetic data often suffer from a domain gap, which limits their ability to generalize to real-world scenes and conditions, especially in unconstrained environments. Recent efforts have targeted real-world datasets to produce ground-truth annotations~\cite{grosse2009ground, bell2014intrinsic, li2018learning, careagaIntrinsic}. These datasets are typically captured under constrained setups and require manual annotation or physically-based labeling methods, such as those used in the MIDIntrinsics dataset~\cite{careagaIntrinsic}, an extension of the Multi-Illumination Dataset~\cite{murmann19}. However, such datasets are expensive to collect and do not capture the full complexity of real-world scenes and in-the-wild images.

\paragraph{Self-supervised intrinsic decomposition.}
Using changes in lighting while keeping the scene static was first explored by~\cite{weiss2001deriving, laffont2015intrinsic, loscos:inria-00527577}, based on the idea that the base color remains constant while illumination varies across time-lapse sequences. Li et al.~\cite{li2018learning} extended this idea by collecting internet time-lapse videos to build the BigTime dataset, training a CNN on sequences of the same scene under different lighting conditions, constrained by consistency and smoothness losses to separate reflectance and shading.
Similarly, Ma et al.~\cite{ma2018single} proposed a two-stream network trained with similar assumptions. While both methods showed strong results at the time, they struggle to generalize to complex real-world lighting scenarios, especially on challenging materials such as glossy or reflective surfaces, where inferring a correct albedo becomes highly ambiguous.
More recently, Zhang et al.~\cite{zhang2024latent} leverages intrinsic components in the latent space and learn to relight scenes from image pairs. Their method can produce realistic relighting results while inferring albedo-like maps that are less affected by lighting conditions. However, the predicted albedo still contains specular highlights, and the model shows limited performance on out-of-domain scenes.

\begin{figure*}[t]
    \centering
    \begin{overpic}[width=0.75\textwidth]{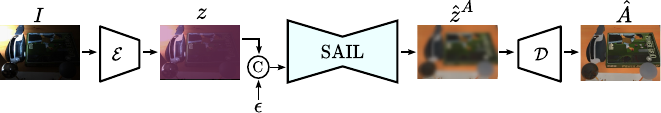}
    \end{overpic}
    \caption{\textbf{\method~ overview -- } Given a single input image, encoded into the latent space using the frozen pre-trained VAE encoder, \method~estimates an albedo representation in latent space, which, when decoded, produces an albedo image without any lighting effects.}
    \label{fig:method-architecture}
    \hfill
\end{figure*}

\paragraph{Generative models for image decomposition.}
Generative models have become popular due to their strong performance in image generation for relighting~\cite{bhattad2022stylitgan} and inferring scene properties~\cite{du2023generative}. Recent work has shown that strong priors learned by diffusion models trained on large-scale datasets can encode meaningful physical and structural information~\cite{su2024compositionalimagedecompositiondiffusion, du2023generative}. Building on this, recent methods~\cite{kocsis2023intrinsic, Zeng_2024} leverage diffusion models for material estimation. However, these approaches rely on ground-truth supervision, such as annotated albedo. In contrast, our approach does not use labeled data.
We instead utilize the prior knowledge from a latent diffusion model to perform unconditioned scene relighting between image pairs, serving as a surrogate objective for albedo estimation on real-world images.

\section{Method}
Our method, dubbed \method~and shown in Fig.~\ref{fig:method-architecture}, estimates, from a single image and without requiring any labeled ground-truth data, a lighting-invariant output free of illumination effects, which we term an \textit{albedo-like} image.
Our approach, detailed in Section~\ref{sec:diffusion}, repurposes the prior knowledge of a latent diffusion model for unconditioned scene relighting.
This acts as surrogate supervision, allowing albedo estimation to be trained with unlabeled multi-illumination image sequences.
In Section~\ref{sec:latent_rep}, we present the core idea of our contribution: a novel approach to intrinsic image decomposition, characterized by its full representation in the latent space allowing extraction of albedo.
Finally, Section~\ref{sec:latent_constraints} describes the proposed regularizations designed to guide the training towards consistent light-invariant properties.

\subsection{Unconditioned Relighting as Surrogate Objective for Albedo Estimation}
\label{sec:diffusion}

Recent advances in diffusion models have demonstrated their effectiveness in learning and decomposing scene properties~\cite{su2024compositionalimagedecompositiondiffusion,du2023generative}.
By progressively denoising a Gaussian noise, these models generate samples that are aligned with the training distribution.
Our goal is to leverage the prior knowledge of a latent diffusion model for albedo estimation.
To achieve this, we adapt a pre-trained Stable Diffusion model~\cite{rombach2022high} by repurposing its objective for unconditioned scene relighting.

Let $\mathcal{E}$ and $\mathcal{D}$ be the frozen pre-trained VAE encoder and decoder, respectively.
Let $I_i$ and $I_j$ be two images of the same scene, taken at the same viewpoint, under different lighting conditions $i$ and $j$, with respective latent representation $z_i$ and $z_j$ computed with $\mathcal{E}$.
To train our model for unconditioned scene relighting, we adopt a more suitable image-to-image model architecture rather than a text-to-image one.
We concatenate the conditioning latent $z_i$ with the noisy latent $z_j + \epsilon$, where $\epsilon$ is a Gaussian noise.
The model is then trained to reconstruct the latent image under the lighting condition $j$, enabling it to learn image intrinsic property without relying on labeled data.

The training objective is to minimize the difference between the original added noise $\epsilon$ and the predicted noise, at a specific timestep $t$, following the $\epsilon\text{-prediction}$ loss:
\begin{equation}
\mathcal{L} = \mathbb{E}_{z_j, \epsilon \sim \mathcal{N}(0, I), t} \left[ \left\| \epsilon - F_\theta(z_j + \epsilon, \, t, \, z_i) \right\|_2^2 \right],
\label{eq:noise-pred}
\end{equation}
with $F_\theta$ being a noise prediction function parametrized by the trainable weights $\theta$ and conditioned on timestep $t$ and reference latent $ z_i$.

\begin{figure*}[t]
    \centering
    \begin{overpic}[width=0.85\textwidth]{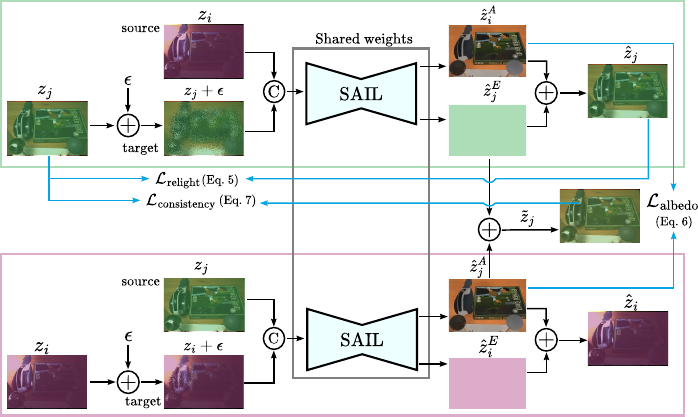}
    \end{overpic}
    \caption{\textbf{\method~training -- } \method~performs image intrinsic decomposition ($\hat{z}_i^A$,  $\hat{z}_i^E$) conditioned on a source latent $z_i$, through a surrogate objective of image relighting  ($\mathcal{L}_{\text{relight}}$ and $\mathcal{L}_{\text{consistency}}$). Considering multiple illuminations $i,j$ of the same scene, we constraint the predicted latent albedo extracted from these sources latents to be identical ($\mathcal{L}_{\text{albedo}}$).
    }
    \label{fig:train-pipeline}
\end{figure*}

\subsection{Albedo estimation}
\label{sec:latent_rep}
\paragraph{Image intrinsic decomposition in the latent space.} Previous approaches~\cite{li2018learning,ma2018single} rely on the classical shading equation, where an image $I_i$ of a specific scene under the lighting condition $i$ is modeled as the product of an albedo $A$ and a shading component $S_i$:

\begin{equation}
I_i = A \times S_i.
\label{eq:shading}
\end{equation}

In contrast with previous work, we propose a novel intrinsic image decomposition fully represented in the latent space:
\begin{equation}
z_i= z^A + z_i^E,
\label{eq:representation}
\end{equation}
where $z^A$ is the latent representation of the albedo image $A$, and $z_i^E$ is a latent component responsible of the lighting effects visible in image $i$.
To recover the albedo in the image space, we decode its latent representation $z^A$ using the VAE decoder $\mathcal{D}$, thus obtaining: $A = \mathcal{D}(z^A)$.
Crucial for our formulation, the latent component $z_i^E$ (Eq.~\ref{eq:representation}) fundamentally differs from the image-domain shading $S_i$ (Eq. ~\ref{eq:shading}). Our lighting representation $z_i^E$ is formulated at a higher level of abstraction within the latent space. We believe that this latent space formulation makes the representation easier for the model to learn, since it is not strictly required to produce a physically plausible shading component for the relighted image.

\paragraph{Compositional latent intrinsic decomposition with diffusion.}
We adapt our diffusion model to directly output the intrinsic image decomposition detailed in the preceding paragraph as follows:
\begin{equation}
    F_\theta(z_j + \epsilon, \, t, \, z_i) = \left\{ \hat{z}_i^A, \hat{z}_j^E \right\},
\end{equation}
where $\hat{z}_i^A$ is the estimated albedo latent predicted by the model conditioned on latent $z_i$, and $\hat{z}_j^E$ is the predicted lighting component of the latent $z_j$. 
The subscript $i$ highlights that the predicted albedo latent $\hat{z}_i^A$ is obtained with the conditioning of the image under lighting condition $i$.
By combining the decomposition predicted by the model, we derive the predicted latent representation of the relighted image using: $\hat{z_j} = \hat{z}_i^A + \hat{z}_j^E$.

Since our new model formulation directly predicts the denoised target latent, the standard noise prediction loss ($\epsilon$-prediction in Eq.~\ref{eq:noise-pred}) is incompatible. We instead adopt a sample prediction strategy, and our relighting loss is thus supervised as:
\begin{equation} \label{eq:relight_loss}
    \mathcal{L}_{\text{relight}} = \left\| z_j - (\hat{z}_i^A + \hat{z}_j^E) \right\|_2^2 =  \left\| z_j - \hat{z_j} \right\|_2^2. 
\end{equation}

We draw inspiration from recent work on compositional image decomposition with diffusion models~\cite{su2024compositionalimagedecompositiondiffusion}, which propose an unsupervised method to separate an image into independent 'object factors'~\cite{du2021unsupervised}.
The authors show that this decomposition can be optimized within a standard diffusion framework by summing multiple outputs of a noise prediction model. In \method, rather than decomposing into 'object factors', we propose an intrinsic image decomposition fully represented in the latent space, where the albedo representation and a light-dependent term are combined to recompose the relighted latent image.

\subsection{Regularizing the latent representation}
\label{sec:latent_constraints}
\paragraph{Latent albedo regularization.}
To encourage consistent predicted light-invariant properties for a given scene under $K$ different illumination conditions, we introduce a self-supervision loss on the predicted latent albedo:
\begin{equation} \label{eq:albedo-loss}
\mathcal{L}_{\text{albedo}} = \sum_{\substack{i,j \in K \\ i \neq j}} \left\| \hat{z}_i^A - \hat{z}_{j}^A \right\|_2^2.
\end{equation}
The self-supervision loss $\mathcal{L}_{\text{albedo}}$ enforces the predicted latent albedos to be consistent across different lighting conditions for the same scene.

As an additional self-supervision constraint, we introduce a cross-consistency term: 
\begin{equation}
\mathcal{L}_{\text{consistency}} = \left\| z_i - (\hat{z}_i^A + \hat{z}_i^E) \right\|_2^2,
\label{eq:consistency}
\end{equation}
where $\hat{z}_i^A$ is the latent albedo predicted by $F_\theta$ under the conditioning of latent $z_i$, and $\hat{z}_i^E$ is the lighting component predicted by $F_\theta$ under the conditioning of a different latent $z_j, j\neq i$. 
This cross-consistency loss enforces the model to reconstruct different lighting conditions with the same latent albedo. 
Although the previously introduced self-supervision constraints provide strong-signal supervision, they are not sufficient on their own. Enforcing the prediction of identical $\hat{z}^A$ across varying lighting conditions still leaves ambiguity in the decomposition. In particular, the model could trivially satisfy this constraint by predicting a constant value, unrelated to the real latent albedo, and backed all information within the latent lighting component.
Therefore, additional constraints and regularization are necessary to ensure that the predicted light-invariant latents encode meaningful scene information rather than collapsing to trivial solutions.

Following the Retinex Theory~\cite{land1971lightness}, large gradients in an image are attributed to albedo properties (such as edges, colors, and textures), while smooth variations are associated with lighting effects. This can be summarized as: ``albedo is more similar to the image than shading''. To encourage this behavior in our model, we constrain the latent representation of illumination-invariant properties to be close to the latent encoding of the input image. This constraint ensures that $\hat{z}^A$  captures meaningful scene information rather than collapsing to a trivial solution. We introduce this constraint as the invariant loss function:
\begin{equation}
    \mathcal{L}_\text{invariant} = \left\| z_j - \hat{z}_i^A \right\|_2^2.
    \label{eq:invariant}
\end{equation}
This regularization term is conflicting with the relighting loss used to train our diffusion model (Eq.~\ref{eq:relight_loss}), as a result, $\hat{z}_i^A$ will never exactly reproduce the relighted latent image but will preserve its high frequency details. 

\paragraph{Latent lighting component regularization.}
Initial results obtained by applying constraints \textbf{only} on the latent albedo component of our model were not satisfactory. Many high frequency details were \textit{leaking} into the latent lighting component, resulting on poorly detailed albedo estimation.

To prevent such behavior, we add two additional constraints on $\hat{z}_i^E$. 
We randomly smooth $\hat{z}_i^E$ using a Gaussian blur spatial filter during training to reduce the impact of high frequency component predicted by the model. We additionally add a regularization loss to match the distribution of the predicted $\hat{z}_i^E$ to an \textit{ideal} latent lighting component distribution. Unfortunately, we do not have access to this \textit{ideal} distribution because, as explained previously, the lighting latent component  $\hat{z}_i^E$ that we are predicting does not correspond to any physically measurable property in the image space. To estimate this distribution, we run an experiment in a fully supervised setup using ground truth albedo from the MIDIntrinsics dataset~~\cite{careagaIntrinsic} and analyze the value of predicted $\hat{z}_i^E$ (see our supplementary material for further details). We found that the distribution of $\hat{z}_i^E$ is largely shifted toward negative values, hence we add a simple yet effective constraint on the model prediction:
\begin{equation}
\mathcal{L}_{\text{reg}} = \max(0, \hat{z}_i^E).
\label{eq:reg}
\end{equation}
$\mathcal{L}_{\text{reg}}$ penalizes positive values in $\hat{z}_i^E$, reducing its expressiveness and leading to better latent intrinsic decomposition.

\subsection{Complete training objective}
An overview of our training pipeline can be found in figure~\ref{fig:train-pipeline}. The final training objective is the combination of the relighting losses, self-supervision terms and regularization constraints:
\begin{equation}
    \mathcal{L} = \mathcal{L_{\text{relight}}} + 
    \mathcal{L_{\text{albedo}}} + 
    \mathcal{L_{\text{consistency}}} +
    \lambda \left( \mathcal{L_{\text{invariant}}} + 
     \mathcal{L_{\text{reg}}} \right),
\end{equation}
where $\lambda$ is a balancing factor for regularization losses, set to $0.5$ for all our experiments.

\subsection{Single-image albedo prediction}
At inference, we follow a standard denoising process described in Algorithm~\ref{alg:inference} to obtain, from a real-world input image, an albedo-like image.

\begin{algorithm}
\caption{Albedo generation with SAIL (inference stage)}\label{alg:inference}
\begin{algorithmic}
\Require Diffusion steps $T$, SAIL model $F_\theta$, conditioning image $I_i$, VAE encoder $\mathcal{E}$ and decoder $\mathcal{D}$
\Ensure Predicted albedo-like image $\hat{A}$
\State $z_i \gets \mathcal{E}(I_i)$
\State $\hat{z}_j \sim \mathcal{N}(0, \mathbf{I})$ \Comment{Initialize the noisy relighted latent from a normal distribution}
\For{$t = T,...,1$}
    \State $\hat{z}^A_i, \hat{z}^E_j \gets F_\theta(\hat{z}_j, t, z_i)$  \Comment{Denoise and decompose the noisy relighted latent}
    \State $\hat{z}_j \gets \hat{z}^A_i + \hat{z}^E_j$ \Comment{Recompose the clean relighted latent}
    \State $\hat{z}_j \gets {\text{DDIM}}(\hat{z}_j, t) $ \Comment{Perform one noise scheduling step}
\EndFor
\State $\hat{A} \gets \mathcal{D}(\hat{z}^A_i)$
\end{algorithmic}
\end{algorithm}

\section{Experiments}

\subsection{Experiments details}

\paragraph{Implementation details.}

We fine-tune a pre-trained Stable Diffusion V2.1~\cite{rombach2022high} with a batch size of 40 for 35k steps using a fixed learning rate of $1\times10^{-5}$ and the Adam optimizer. Training is performed with random $512\times512$ crops, and evaluation is done on resized images preserving the original aspect ratio with a minimum height or width of 512. The model is trained for approximately 6 days on 4 GPUs equivalent to Nvidia H200. 
At inference, we use DDIM sampling~\cite{song2020denoising} with 50 steps and classifier-free guidance~\cite{ho2022classifierfreediffusionguidance} applied to the conditioning image latent with a guidance scale of 1.5. 
Following~\cite{kocsis2023intrinsic}, we sample 10 albedo latents per image, compute their mean, and evaluate metrics on the resulting averaged prediction.

\paragraph{Datasets.}
We train our model on the MIT Dataset~\cite{murmann19}, which consists of 985 scenes captured under 25 different illumination conditions, following the original data split. Additionally, we augment the dataset with 150 indoor scenes from the BigTime dataset~\cite{li2018learning}, randomly splitting them into 142 scenes for training and 12 scenes for validation.

\paragraph{Baselines.}
We compare our method with two self-supervised methods~\cite{li2018learning, zhang2024latent} and two recent supervised diffusion-based methods~\cite{kocsis2023intrinsic, Zeng_2024}. Due to the older implementation of~\cite{li2018learning}, we only report their results on the IIW dataset, as they publicly provide predictions on their website. A comparison of each method features can be found in Tab~\ref{tab:method_comparison}.

\paragraph{Albedo consistency metrics.} 

To evaluate the consistency of the predicted albedo under varying illumination conditions, we report PSNR and SSIM metrics using the ground-truth albedo from~\cite{careagaIntrinsic} on the Multi-Illumination Dataset~\cite{murmann19}. For each scene, we compute the average metric over 25 illuminations, and report the mean metrics across all scenes.

 \paragraph{Real-world albedo evaluation metrics.} 
    Due to the lack of ground-truth albedo annotations, IIW~\cite{bell2014intrinsic} and MAW~\cite{wu2023measured} were introduced to evaluate in the wild albedo predictions. IIW collects human-based relative brightness judgments and proposes the WHDR metric, though it does not always reflect the qualitative quality of the predicted albedo~\cite{forsyth2021intrinsic, wu2023measured}. MAW addresses these limitations with denser annotations and new metrics evaluating intensity, chromaticity, and texture errors. We report WHDR on IIW, and WHDR, intensity, chromaticity, and texture on MAW. All WHDR results on~\cite{bell2014intrinsic} are recomputed on the subset of images available from~\cite{li2018learning}; thus, reported metrics may differ from original papers.

\subsection{Quantitative results}
\label{sec:quantitative}

\paragraph{Consistency to diverse lighting condition.}
We report in Tab.~\ref{tab:mid_eval} the consistency evaluation results on the MIDIntrinsics dataset~\cite{careagaIntrinsic}. \method~outperforms both supervised and self-supervised baselines, demonstrating greater robustness to complex real-world lighting conditions. Supervised methods~\cite{kocsis2023intrinsic,Zeng_2024}, trained on synthetic datasets with idealized lighting, show lower performance, confirming their limited generalization to real-world conditions.

\paragraph{In the wild images evaluation.}
We report in Tab.~\ref{tab:maw} the WHDR, intensity, chromaticity, and texture metrics evaluated on the IIW~\cite{bell2014intrinsic} and MAW~\cite{wu2023measured} datasets.
\method~achieves comparable results to state-of-the-art methods on WHDR for both datasets. 
Regarding intensity, chromaticity and texture errors, \method~achieves comparable results with existing state-of-the-art solutions.
$\text{RGB} \to\text{X}$\cite{Zeng_2024} and IntrinsicDiffusion~\cite{kocsis2023intrinsic} report comparable results to self-supervised approaches, as the test datasets~\cite{bell2014intrinsic,wu2023measured} mostly fall within their training domain (indoor images captured under idealized lighting conditions). 
We draw the reader's attention to the fact that the metrics reported in Tab.~\ref{tab:maw} are computed based on available annotations in the MAW dataset, which correspond to small, planar, and homogeneous patches (e.g., sofas, pillows) and do not capture specular objects, reflections, shadows, or the overall color distribution of the predicted albedos. This remains a limitation for quantitatively evaluating in-the-wild albedo predictions.

\begin{table}[h!]
\centering
\begin{minipage}{0.55\linewidth}
\centering
\resizebox{0.98\linewidth}{!}{%
\begin{tabular}{lcccc}
\toprule
\textbf{Method} & \textbf{Self-supervised} & \textbf{Diffusion} & \textbf{Data} & \textbf{Prediction} \\
\midrule
Li et al.~\cite{li2018learning} & \checkmark & \ding{55} & Real & Albedo+Shading\\
IntrinsicDiffusion.~\cite{kocsis2023intrinsic} & \ding{55} & \checkmark & Synth & Albedo+BRDF\\
$\text{RGB} \to \text{X}$~\cite{Zeng_2024} & \ding{55} & \checkmark & Synth & Albedo+BRDF\\
LatentIntrinsics.~\cite{zhang2024latent} & \checkmark & \ding{55} & Real & Relighting\\
\method~(Ours) & \checkmark & \checkmark & Real & Albedo\\
\bottomrule
\end{tabular}}
\caption{\footnotesize Comparison of baselines and our method in terms of albedo supervision, method type and training data type.}
\label{tab:method_comparison}
\end{minipage}%
\hfill
\begin{minipage}{0.4\linewidth}
\centering
\resizebox{0.7\linewidth}{!}{%
\begin{tabular}{lcc}
\toprule
\textbf{Method} & \textbf{PSNR} & \textbf{SSIM} \\
\midrule
IntrinsicDiffusion~\cite{kocsis2023intrinsic} & 11.73    & 0.39   \\
$\text{RGB} \to\text{X}$~\cite{Zeng_2024}             & 7.19  & 0.35 \\
\midrule
LatentIntrinsics~\cite{zhang2024latent}      &  13.12 &  0.54 \\
\method (Ours)                           &  \textbf{17.00} &  \textbf{0.63} \\
\bottomrule
\end{tabular}}
\caption{\footnotesize Quantitative results on the MIDIntrinsics dataset~\cite{careagaIntrinsic}. We report the mean PSNR and SSIM over 30 test scenes.}
\label{tab:mid_eval}
\end{minipage}
\end{table}

\vspace{-5mm}
\begin{table}[h!]
\centering
\resizebox{0.65\linewidth}{!}{%
\begin{tabular}{lc|cccc}
\toprule
\multirow{2}{*}{\textbf{Method}} & IIW dataset~\cite{bell2014intrinsic} & \multicolumn{4}{c}{MAW dataset~\cite{wu2023measured}} \\
 & \textbf{WHDR} & \textbf{WHDR} & \textbf{Intensity} & \textbf{Chromaticity} & \textbf{Texture}\\
\midrule
IntrinsicDiff.~\cite{kocsis2023intrinsic} & 31\% & 44\% & 0.014 & 5.04 & 0.4413 \\
$\text{RGB} \to \text{X}$~\cite{Zeng_2024} & 21\% & 22\% & 0.007 & 3.81 & 0.4317 \\
\midrule
Li et al.~\cite{li2018learning} & 36\% & -- & -- & -- & -- \\
LatentIntr.~\cite{zhang2024latent} & 44\% & \textbf{27\%} & \textbf{0.010} & 5.82 & 0.4683 \\
\method~(Ours) & \textbf{34\%} & 33\% & \textbf{0.010} & \textbf{5.71} & \textbf{0.4549} \\
\bottomrule
\end{tabular}}
\caption{\footnotesize Quantitative results on IIW dataset~\cite{bell2014intrinsic} and MAW dataset~\cite{wu2023measured}. We report the WHDR, Intensity, Chromacity and Texture errors.}
\label{tab:maw}
\end{table}
\vspace{-5mm}

\subsection{Qualitative results}

\begin{figure*}[tb] 
\centering
\def\fgsize{0.48}
\def\rowspacing{0.2cm}
\scriptsize
\setlength{\tabcolsep}{0.0035\linewidth}
\renewcommand{\arraystretch}{1.0}
\begin{tabular}{cccccc}%
      Input & IntrinsicDiffusion~\cite{kocsis2023intrinsic} &$\text{RGB} \to\text{X}$~\cite{Zeng_2024} & LatentIntrinsics~\cite{zhang2024latent} & \method~(ours) & GT \\ 

 \vspace{\rowspacing}
    \includegraphics[clip=false, trim={0 0 0 0},width=0.15\columnwidth]{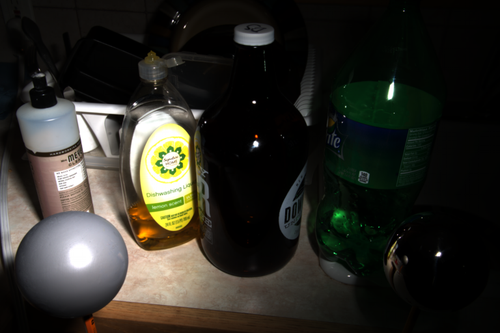} & 
        \includegraphics[clip=false, trim={0 0 0 0},width=0.15\columnwidth]{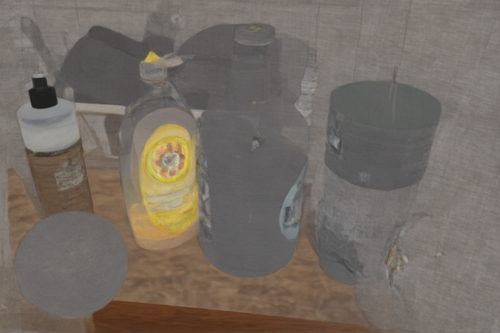} & 
    \includegraphics[clip=false, trim={0 0 0 0},,width=0.15\columnwidth]{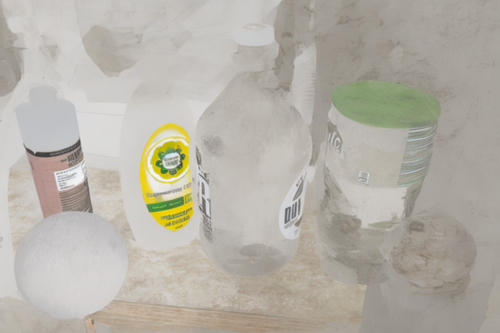} & 
    \includegraphics[clip=false, trim={0 0 0 0},,width=0.15\columnwidth]{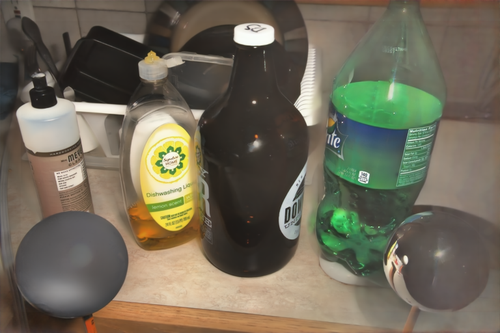} & 
    \includegraphics[clip=false, trim={0 0 0 0},,width=0.15\columnwidth]{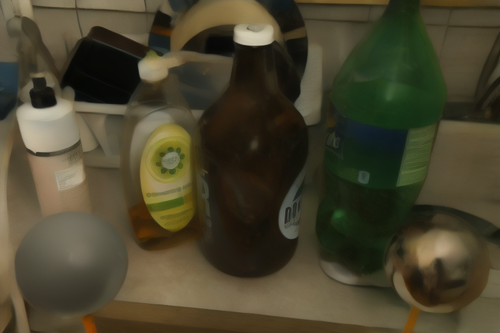} & \\
    \includegraphics[clip=false, trim={0 0 0 0},width=0.15\columnwidth]{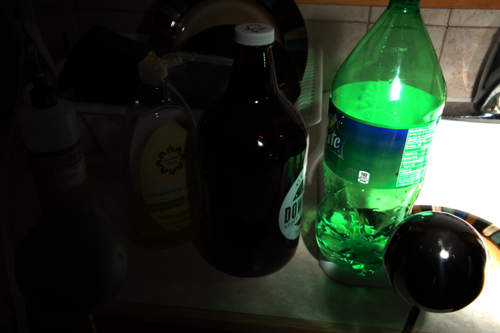} & 
        \includegraphics[clip=false, trim={0 0 0 0},width=0.15\columnwidth]{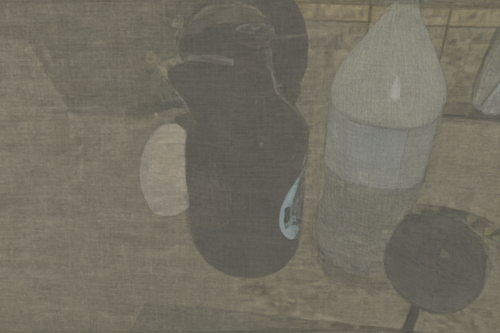} & 
    \includegraphics[clip=false, trim={0 0 0 0},,width=0.15\columnwidth]{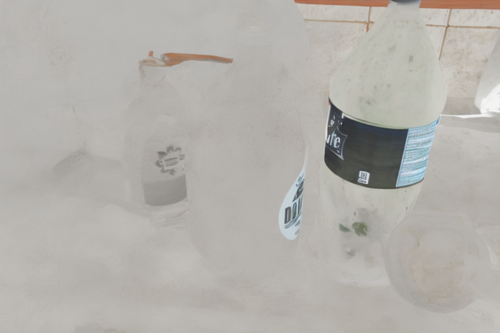} & 
    \includegraphics[clip=false, trim={0 0 0 0},,width=0.15\columnwidth]{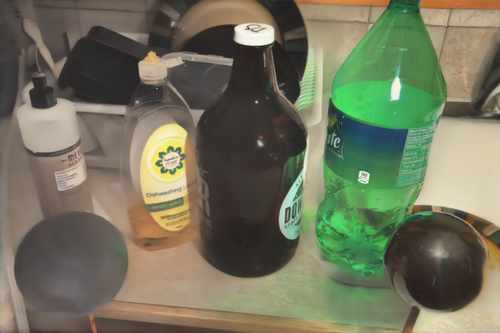} & 
    \includegraphics[clip=false, trim={0 0 0 0},,width=0.15\columnwidth]{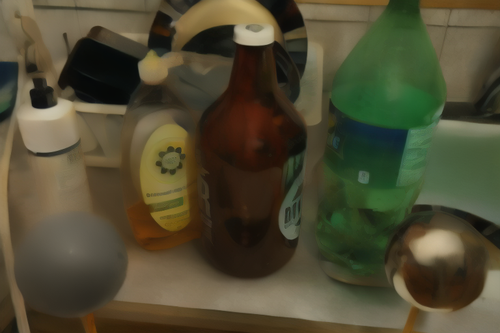} 
     & 
    \includegraphics[clip=false, trim={0 0 0 0},,width=0.15\columnwidth]{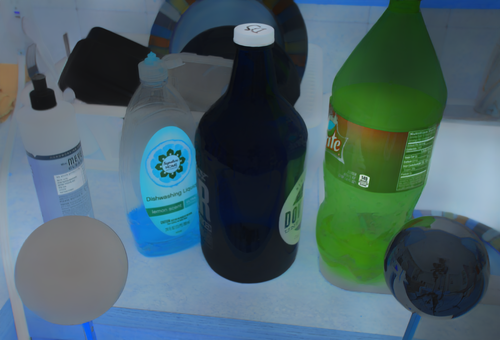}\\
\end{tabular}
\caption{We qualitatively compare the predicted albedos on the MITDataset~\cite{murmann19}. We show that~\method~predicts consistent albedos from the same scene under various lighting conditions.}
\label{fig:middataset}
\vspace{-5mm}
\end{figure*}

\begin{figure*}[tb] 

\centering

\def\fgsize{0.48}
\def\rowspacing{0.2cm}

\scriptsize
\setlength{\tabcolsep}{0.0035\linewidth}
\renewcommand{\arraystretch}{1.0}
\begin{tabular}{ccccc}%

      Input  & IntrinsicDiffusion~\cite{kocsis2023intrinsic}  &$\text{RGB}  \to\text{X}$~\cite{Zeng_2024}&  LatentIntrinsics~\cite{zhang2024latent} &\method~(ours)\\

 \vspace{\rowspacing}

    \includegraphics[clip=false, trim={0 0 0 0},width=0.19\columnwidth]{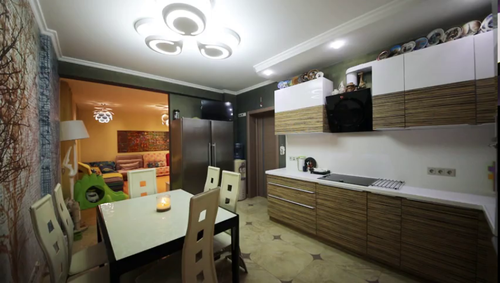} & 
    \includegraphics[clip=false, trim={0 0 0 0},width=0.19\columnwidth]{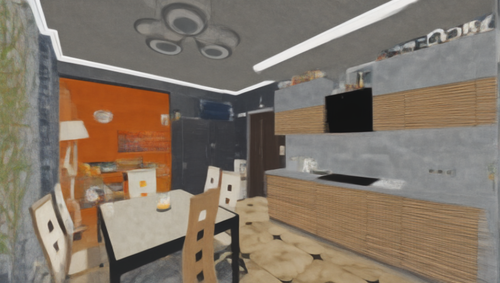} & 
    \includegraphics[clip=false, trim={0 0 0 0},width=0.19\columnwidth]{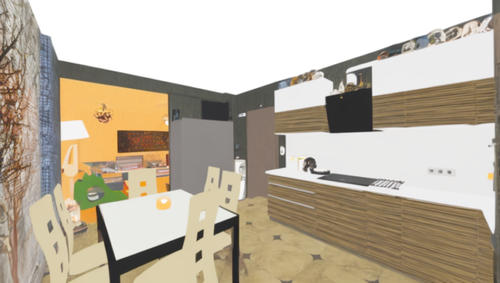} & 
    \includegraphics[clip=false, trim={0 0 0 0},width=0.19\columnwidth]{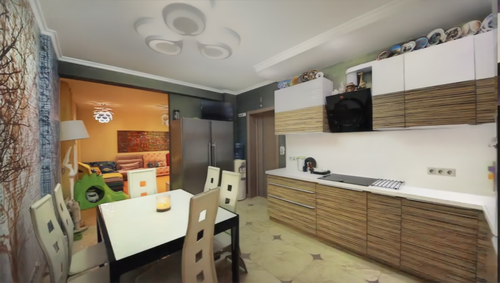} & 
    \includegraphics[clip=false, trim={0 0 0 0},width=0.19\columnwidth]{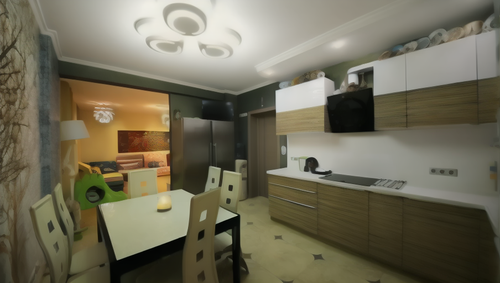} \\

    \includegraphics[clip=false, trim={0 0 0 0},width=0.19\columnwidth]{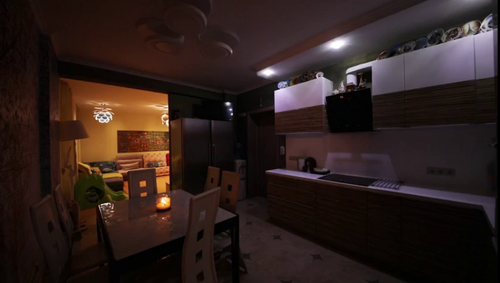} & 
    \includegraphics[clip=false, trim={0 0 0 0},width=0.19\columnwidth]{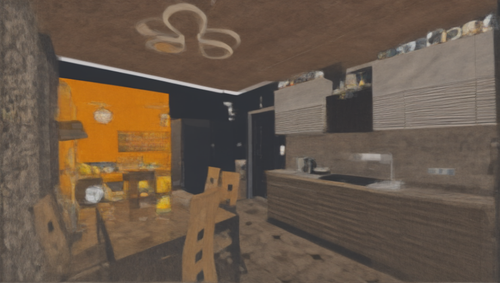} & 
    \includegraphics[clip=false, trim={0 0 0 0},width=0.19\columnwidth]{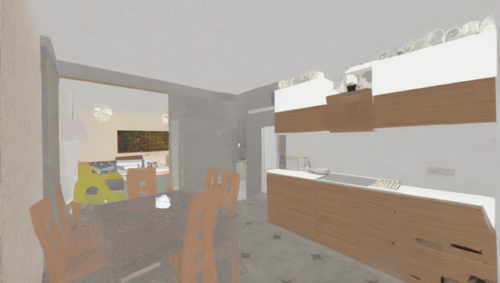} & 
    \includegraphics[clip=false, trim={0 0 0 0},width=0.19\columnwidth]{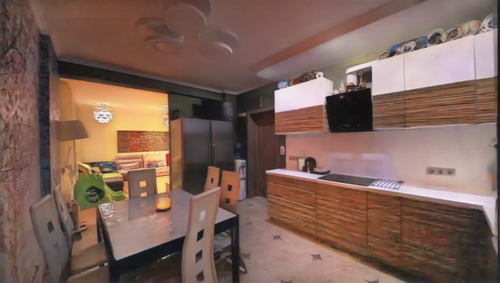} & 
    \includegraphics[clip=false, trim={0 0 0 0},width=0.19\columnwidth]{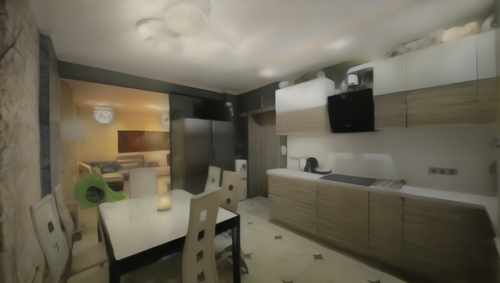} \\

    \includegraphics[clip=false, trim={0 0 0 0},width=0.19\columnwidth]{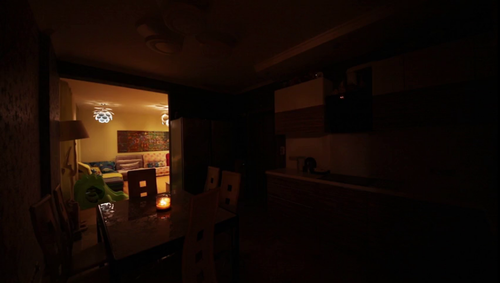} & 
    \includegraphics[clip=false, trim={0 0 0 0},width=0.19\columnwidth]{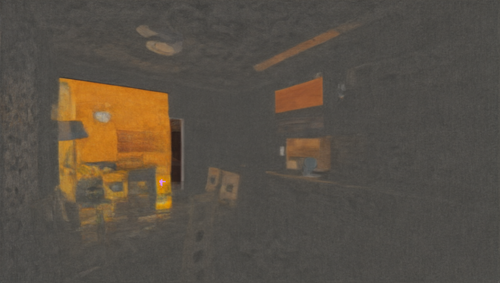} & 
    \includegraphics[clip=false, trim={0 0 0 0},width=0.19\columnwidth]{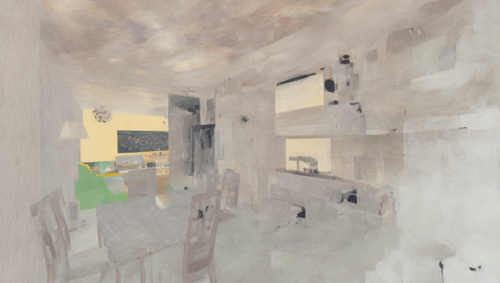} & 
    \includegraphics[clip=false, trim={0 0 0 0},width=0.19\columnwidth]{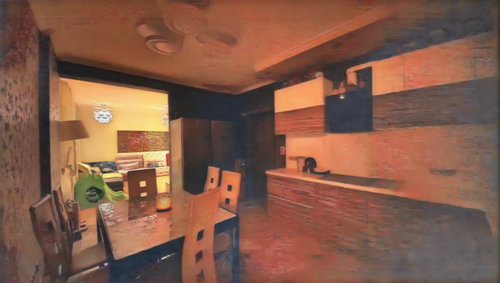} & 
    \includegraphics[clip=false, trim={0 0 0 0},width=0.19\columnwidth]{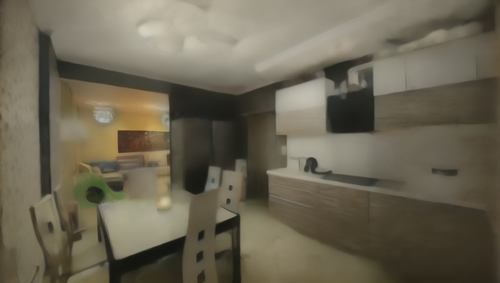} \\
    \midrule
        \includegraphics[clip=false, trim={0 0 0 0},width=0.19\columnwidth]{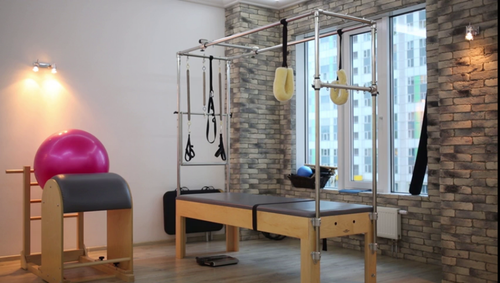} & 
    \includegraphics[clip=false, trim={0 0 0 0},width=0.19\columnwidth]{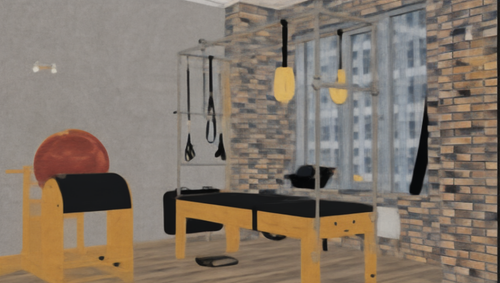} & 
    \includegraphics[clip=false, trim={0 0 0 0},width=0.19\columnwidth]{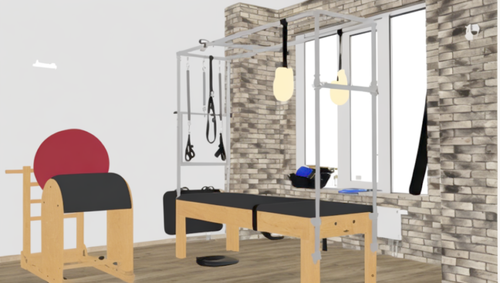} & 
    \includegraphics[clip=false, trim={0 0 0 0},width=0.19\columnwidth]{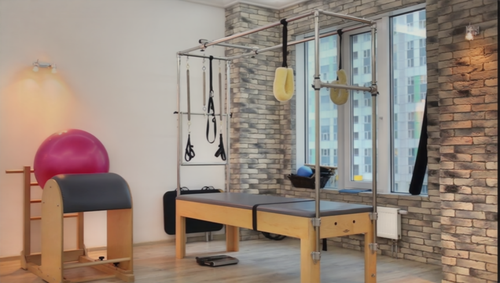} & 
    \includegraphics[clip=false, trim={0 0 0 0},width=0.19\columnwidth]{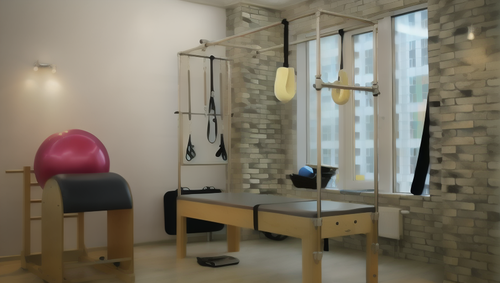} \\

    \includegraphics[clip=false, trim={0 0 0 0},width=0.19\columnwidth]{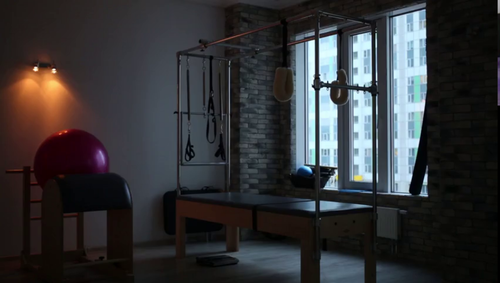} & 
    \includegraphics[clip=false, trim={0 0 0 0},width=0.19\columnwidth]{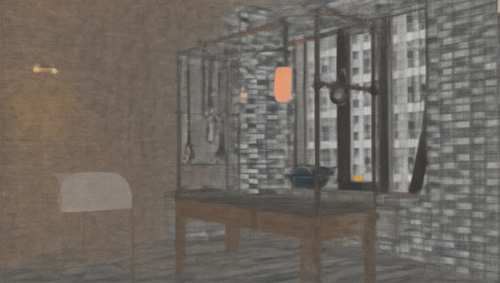} & 
    \includegraphics[clip=false, trim={0 0 0 0},width=0.19\columnwidth]{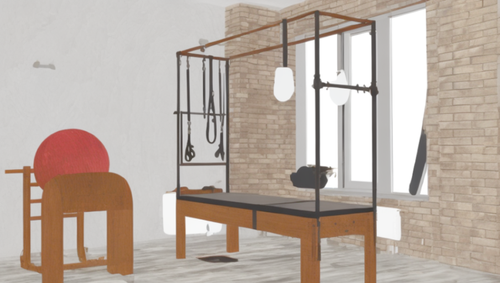} & 
    \includegraphics[clip=false, trim={0 0 0 0},width=0.19\columnwidth]{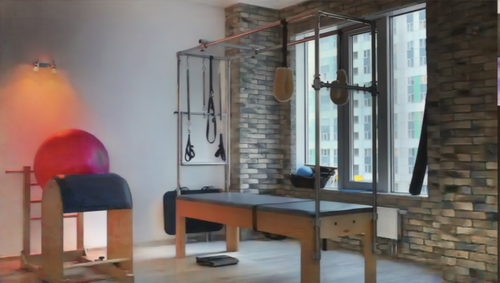} & 
    \includegraphics[clip=false, trim={0 0 0 0},width=0.19\columnwidth]{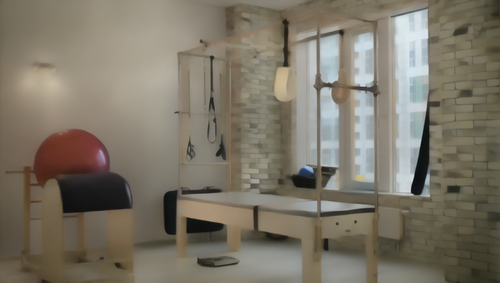} \\

        \includegraphics[clip=false, trim={0 0 0 0},width=0.19\columnwidth]{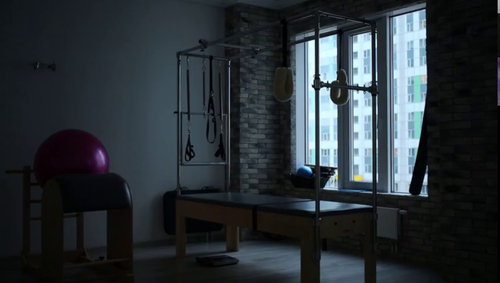} & 
    \includegraphics[clip=false, trim={0 0 0 0},width=0.19\columnwidth]{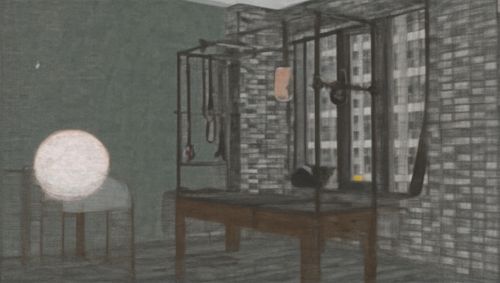} & 
    \includegraphics[clip=false, trim={0 0 0 0},width=0.19\columnwidth]{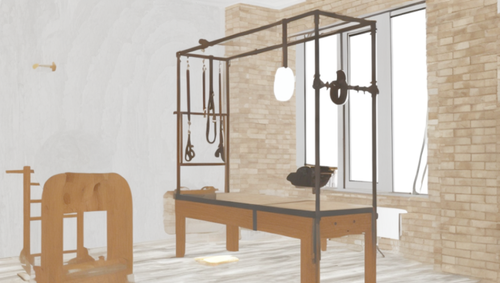} & 
    \includegraphics[clip=false, trim={0 0 0 0},width=0.19\columnwidth]{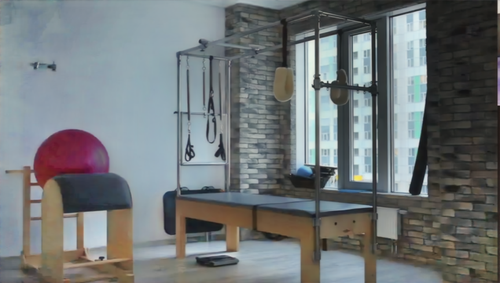} & 
    \includegraphics[clip=false, trim={0 0 0 0},width=0.19\columnwidth]{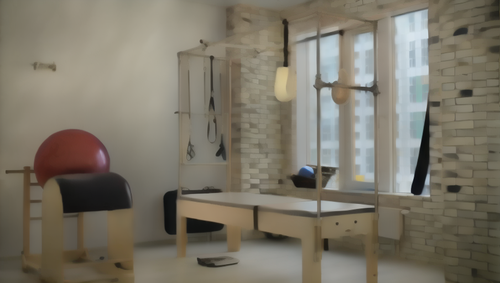} \\

\end{tabular}

  \caption{We qualitatively compare the predicted albedos on the BigTime~\cite{li2018learning}. We show that~\method~predicts consistent albedos from the same scene under various lighting conditions.}
\label{fig:bigtime}

\end{figure*}

The quantitative evaluation in Sec.~\ref{sec:quantitative} relies on human annotations and metrics assessing small planar patches, due to the lack of full ground-truth for in-the-wild images. To complete the analysis, we present qualitative results in Figs.~\ref{fig:middataset}, and~\ref{fig:bigtime}. Results show that \method~predicts consistent albedos across diverse real-world images under complex lighting. IntrinsicDiffusion~\cite{kocsis2023intrinsic} and $\text{RGB} \to\text{X}$~\cite{Zeng_2024} fail to remove lighting effects, producing distortions in object color and appearance. LatentIntrinsics~\cite{zhang2024latent} recovers better albedos than supervised methods but still suffers from reflections and color shifts, especially on glossy and transparent objects. In contrast, \method~successfully separates illumination-variant and invariant properties, even in these challenging cases.

\subsection{Ablation study}
We report in Tab.~\ref{tab:ablations} quantitative and qualitative results of our ablation study conducted on a synthetic Toy Dataset generated with Blender~\cite{blender}. This dataset is composed of 20 different scenes with varying objects, colors, and positions. We evaluate the impact of three key components: latent regularization terms $\mathcal{L}_{\text{invariant}}$ and $\mathcal{L}_{\text{reg}}$ (introduced in Eq.~\ref{eq:invariant} and \ref{eq:reg}), the cross-consistency loss $\mathcal{L}_{\text{consistency}}$ (introduced in Eq.~\ref{eq:consistency}), and the Gaussian blurring of the latent lighting component. The removal of the regularization terms leads the model to collapse to trivial solutions without meaningful albedo structure and colors. Removing the cross-consistency loss degrades prediction stability across illuminations of the same scene.
The partial Gaussian blurring of $\hat{z}_i^E$ improves sharpness and detail preservation in the predicted albedo. 

\begin{table}
    \centering
     \resizebox{0.7\columnwidth}{!}{%
    \begin{tabular}{@{}l*{5}{c}@{}}

    \includegraphics[width=0.20\linewidth]{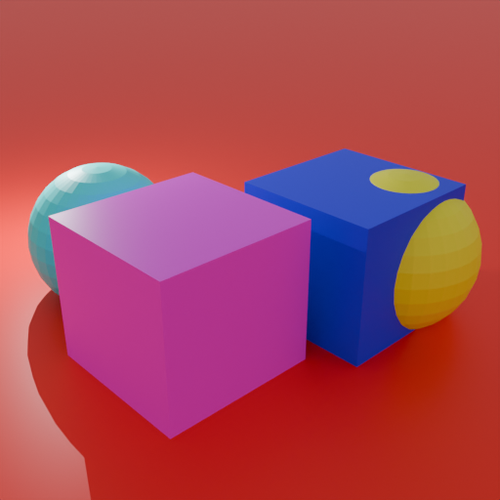}
    & \includegraphics[width=0.20\linewidth]{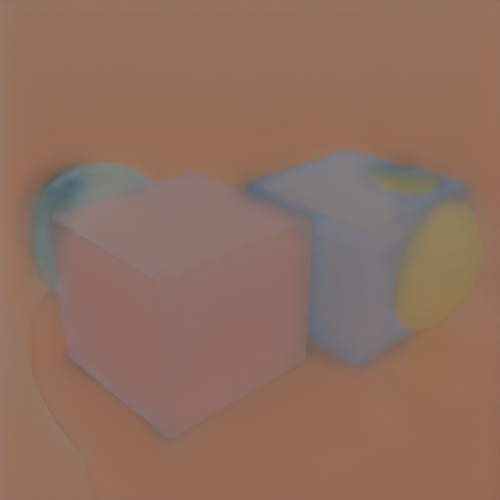}
        & \includegraphics[width=0.20\linewidth]{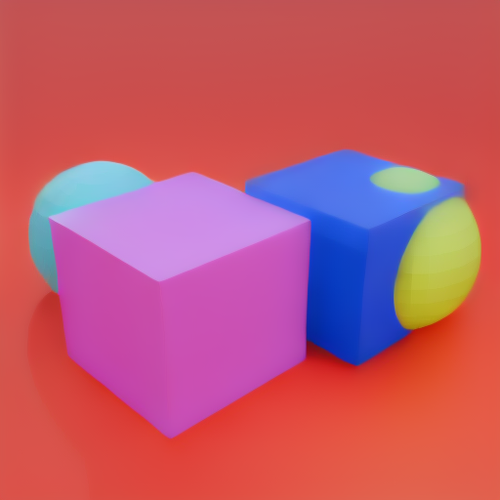}
    & \includegraphics[width=0.20\linewidth]{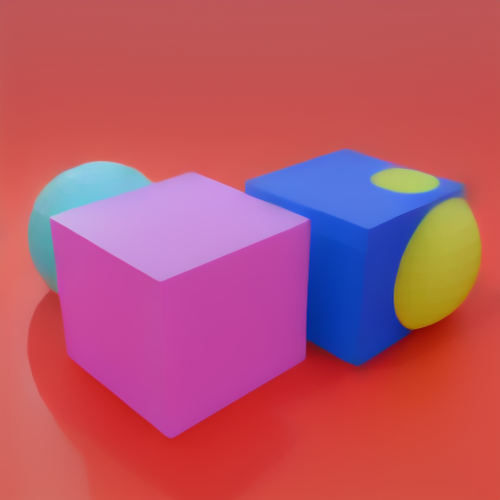}
    
    & \includegraphics[width=0.20\linewidth]{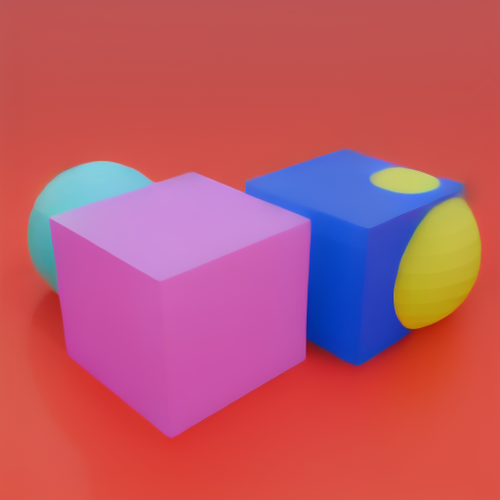}
    \\
        ~~~~Input image & (a) w/o reg. losses & (b) w/o $\mathcal{L}_{\text{consistency}}$ & (b) w/o $\hat{z}_i^E$ blurring & (c) \method~(Ours) \\
        \toprule
        ~~~~PSNR & 15.26 & 16.68 & 17.85  & \textbf{18.15}\\
        ~~~~SSIM & 0.86  & 0.90 & 0.91 & \textbf{0.92 } \\
        \bottomrule
    \end{tabular}}
\caption{ \footnotesize We deactivate key components of \method~to evaluate their impact on albedo estimation. Regularization losses deactivated in (a) are $\mathcal{L}_{\text{invariant}}$ and $\mathcal{L}_{\text{reg}}$.}
\label{tab:ablations}
\end{table}

\section{Discussion}

\paragraph{Applications.}
\method~enables applications such as appearance editing and virtual scene relighting for data augmentation. By predicting consistent albedo representations from a single real-world image, \method~provide valuable data for controlled image relighting with Blender~\cite{blender}, as shown in Fig.~\ref{fig:teaser}.

\paragraph{Limitations.}

While \method~is robust to real-world lighting conditions, its performance depends on the quality and diversity of training data. Some lighting artifacts and shadows remain in predictions. Although more data would improve results, acquiring multiple illuminations of the same scene is challenging, and available time-lapse sequences often lack sufficient lighting variation.

\paragraph{Conclusion.}
In this work, we presented \method, a self-supervised approach to estimate albedo-like representations from unlabeled real-world images. Our method introduces a novel training strategy that repurposes a latent diffusion model for unconditioned scene relighting, using it as a surrogate supervision for albedo prediction. We also proposed a new intrinsic image decomposition fully formulated in the latent space, enabling direct albedo extraction. Quantitative and qualitative evaluations demonstrate that \method~achieves more accurate and consistent albedo predictions across various real-world illumination conditions compared to state-of-the-art methods.

\appendix

\section{Latent lighting component regularization}
\begin{figure*}[tb]
\centering
    \includegraphics[width=0.65\columnwidth]{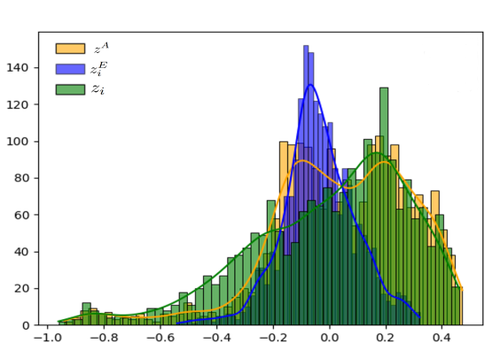}
    \caption{Analysis of of predicted latent distribution}
    \label{fig:latents_distribution}
\end{figure*}

\paragraph{Analysis of Latents Distribution.}
To better understand the statistical behavior of the predicted lighting latent $\hat{z}_i^E$, we conduct an analysis under a fully supervised setup using ground-truth albedo from the MIDIntrinsics dataset~\cite{careagaIntrinsic}. We subtract the predicted albedo $\hat{z}^A$ from the encoded image $z_i$ to isolate the illumination component $\hat{z}_i^E$. Fig.\ref{fig:latents_distribution} shows the distributions of $\hat{z}_i^E$, $\hat{z}^A$, and  $z_i$ across the dataset. We observe that the illumination component $\hat{z}_i^E$ is strongly biased toward negative values, justifying our design choice of applying a non-positivity constraint (see Eq. 9) to ensure that lighting-dependent information does not corrupt the albedo prediction.

\section{Additional results}

\begin{figure*}[tb] 

\centering

\def\fgsize{0.48}
\def\rowspacing{0.2cm}

\scriptsize
\setlength{\tabcolsep}{0.004\linewidth}
\renewcommand{\arraystretch}{1.0}
\begin{tabular}{ccccc}%

      Input  & IntrinsicDiffusion~\cite{kocsis2023intrinsic}  &$\text{RGB}  \to\text{X}$~\cite{Zeng_2024}&  LatentIntrinsics~\cite{zhang2024latent} &\method~(ours)\\

 \vspace{\rowspacing}

    \includegraphics[clip=false, trim={0 0 0 0},width=0.19\columnwidth]{supps/input/33680.png} & 
    \includegraphics[clip=false, trim={0 0 0 0},width=0.19\columnwidth]{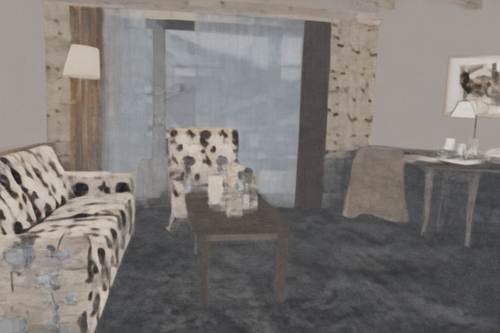} & 
    \includegraphics[clip=false, trim={0 0 0 0},width=0.19\columnwidth]{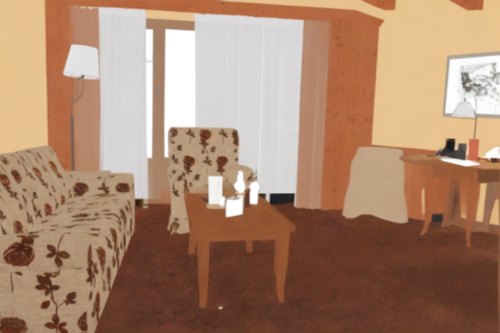} & 
    \includegraphics[clip=false, trim={0 0 0 0},width=0.19\columnwidth]{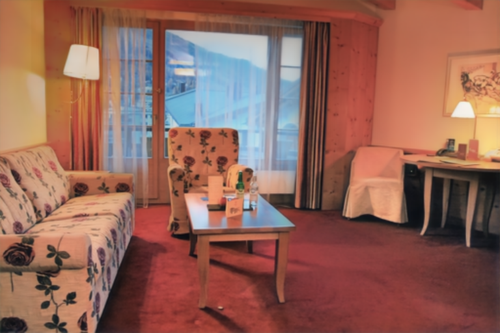} & 
    \includegraphics[clip=false, trim={0 0 0 0},width=0.19\columnwidth]{supps/ours/albedo_mean_33680.png} \\

    \includegraphics[clip=false, trim={0 0 0 0},width=0.19\columnwidth]{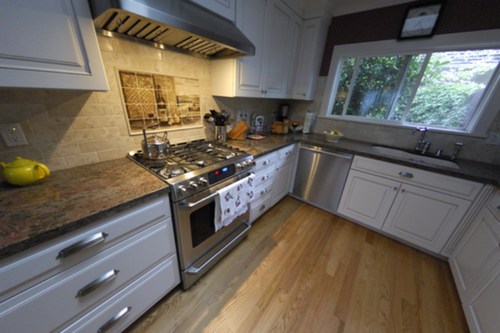} & 
    \includegraphics[clip=false, trim={0 0 0 0},width=0.19\columnwidth]{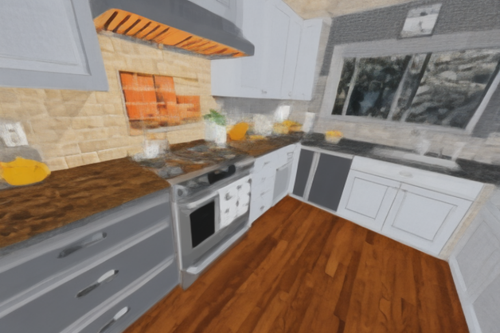} & 
    \includegraphics[clip=false, trim={0 0 0 0},width=0.19\columnwidth]{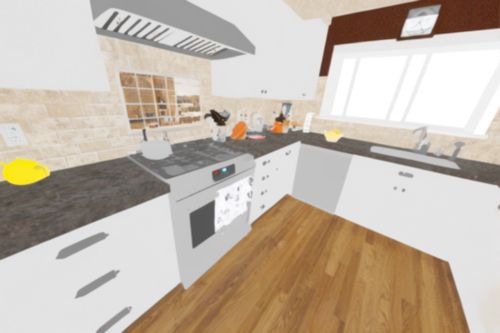} & 
    \includegraphics[clip=false, trim={0 0 0 0},width=0.19\columnwidth]{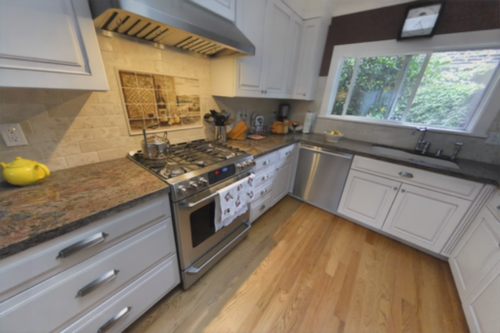} & 
    \includegraphics[clip=false, trim={0 0 0 0},width=0.19\columnwidth]{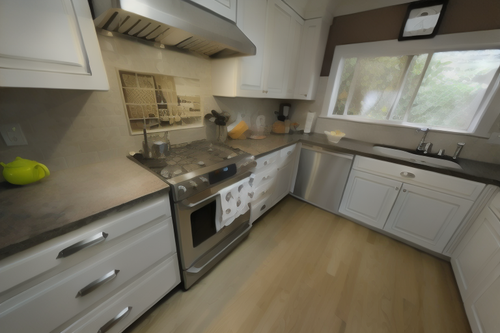} \\

    \includegraphics[clip=false, trim={0 0 0 0},width=0.19\columnwidth]{supps/input/10483.png} & 
    \includegraphics[clip=false, trim={0 0 0 0},width=0.19\columnwidth]{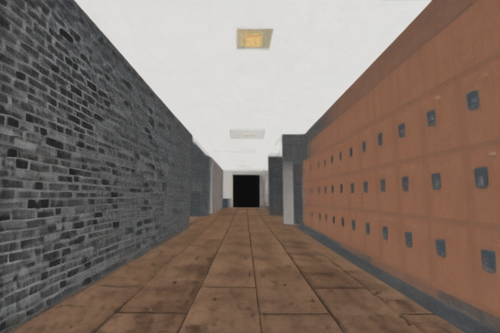} & 
    \includegraphics[clip=false, trim={0 0 0 0},width=0.19\columnwidth]{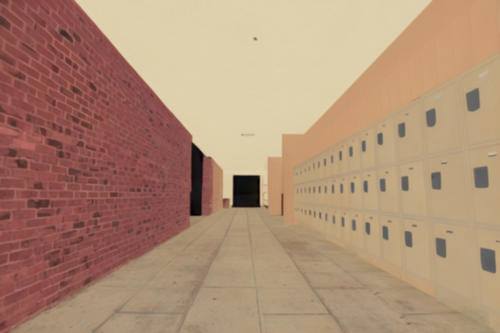} & 
    \includegraphics[clip=false, trim={0 0 0 0},width=0.19\columnwidth]{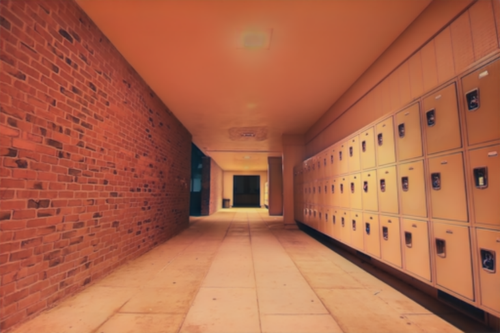} & 
    \includegraphics[clip=false, trim={0 0 0 0},width=0.19\columnwidth]{supps/ours/albedo_mean_10483.png} \\

    \includegraphics[clip=false, trim={0 0 0 0},width=0.19\columnwidth]{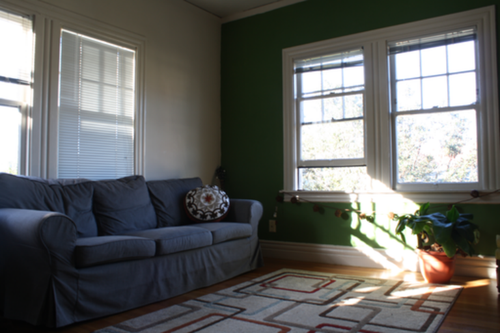} & 
    \includegraphics[clip=false, trim={0 0 0 0},width=0.19\columnwidth]{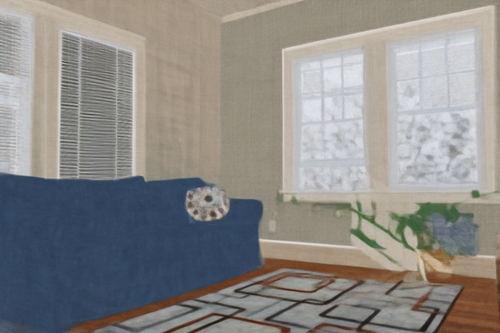} & 
    \includegraphics[clip=false, trim={0 0 0 0},width=0.19\columnwidth]{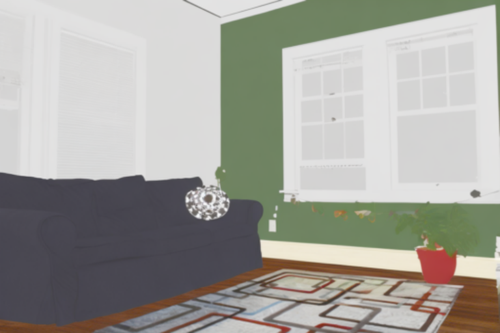} & 
    \includegraphics[clip=false, trim={0 0 0 0},width=0.19\columnwidth]{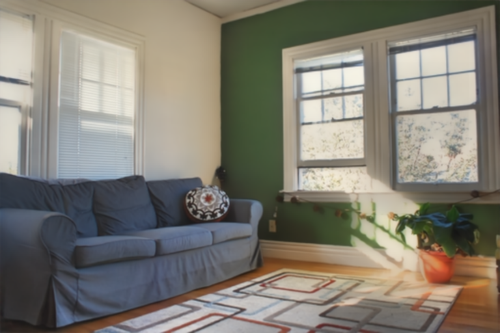} & 
    \includegraphics[clip=false, trim={0 0 0 0},width=0.19\columnwidth]{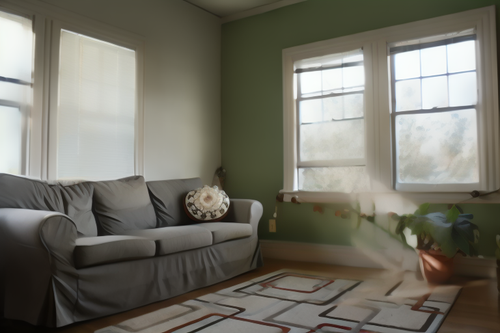} \\
    \midrule

        \includegraphics[clip=false, trim={0 0 0 0},width=0.19\columnwidth]{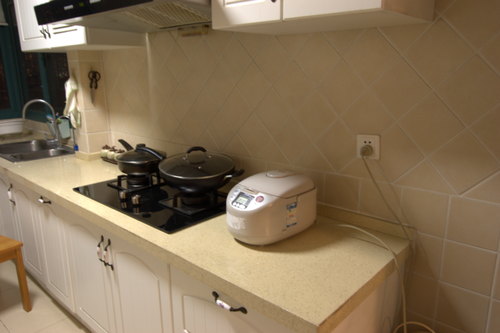} & 
    \includegraphics[clip=false, trim={0 0 0 0},width=0.19\columnwidth]{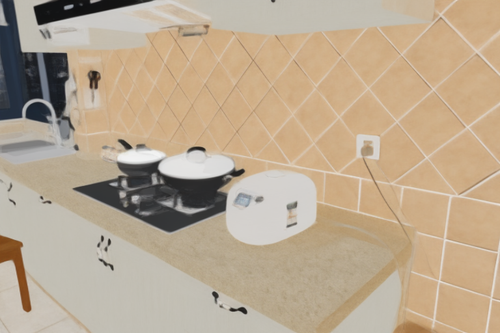} & 
    \includegraphics[clip=false, trim={0 0 0 0},width=0.19\columnwidth]{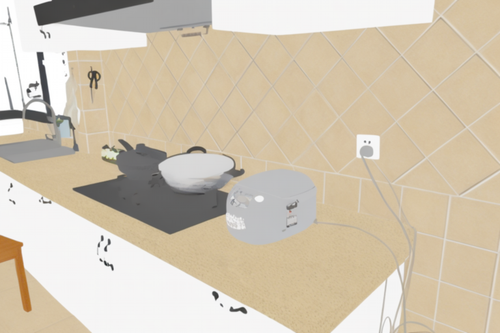} & 
    \includegraphics[clip=false, trim={0 0 0 0},width=0.19\columnwidth]{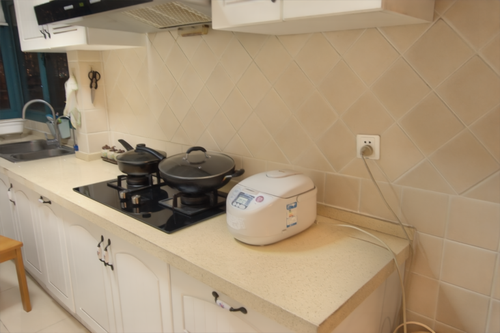} & 
    \includegraphics[clip=false, trim={0 0 0 0},width=0.19\columnwidth]{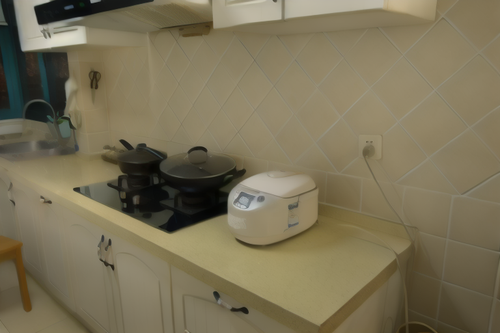} \\

    \includegraphics[clip=false, trim={0 0 0 0},width=0.19\columnwidth]{supps/MAW/INPUT/scene_19/DSC_7002.png} & 
    \includegraphics[clip=false, trim={0 0 0 0},width=0.19\columnwidth]{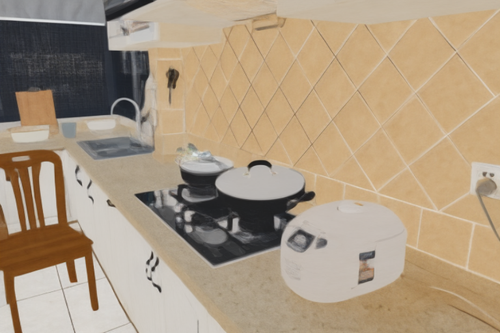} & 
    \includegraphics[clip=false, trim={0 0 0 0},width=0.19\columnwidth]{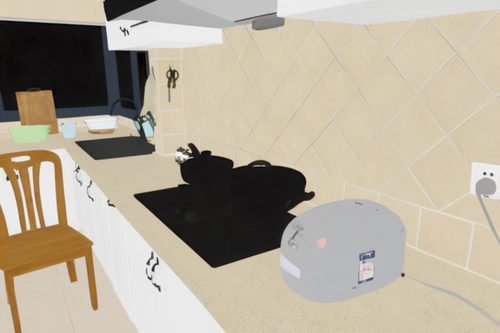} & 
    \includegraphics[clip=false, trim={0 0 0 0},width=0.19\columnwidth]{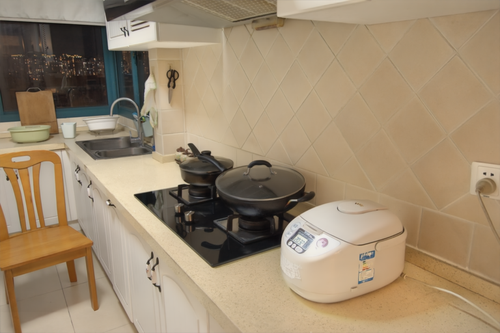} & 
    \includegraphics[clip=false, trim={0 0 0 0},width=0.19\columnwidth]{supps/MAW/ours/scene_19/albedo_mean_DSC_7002.png} \\

    \includegraphics[clip=false, trim={0 0 0 0},width=0.19\columnwidth]{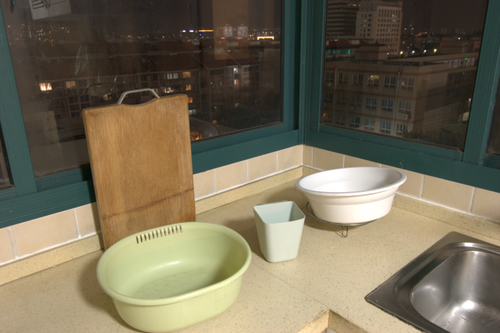} & 
    \includegraphics[clip=false, trim={0 0 0 0},width=0.19\columnwidth]{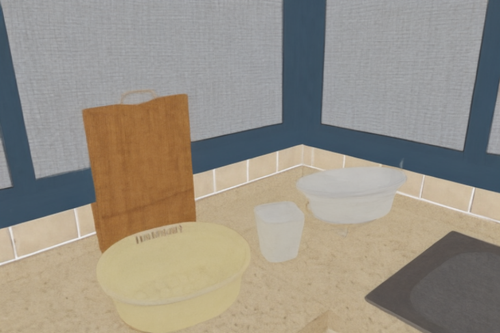} & 
    \includegraphics[clip=false, trim={0 0 0 0},width=0.19\columnwidth]{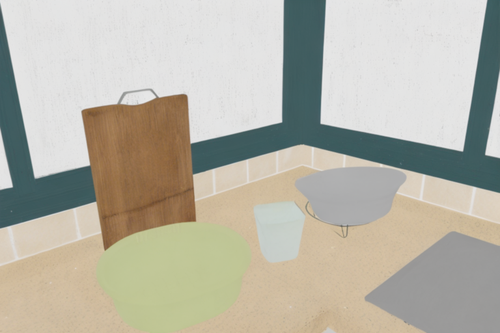} & 
    \includegraphics[clip=false, trim={0 0 0 0},width=0.19\columnwidth]{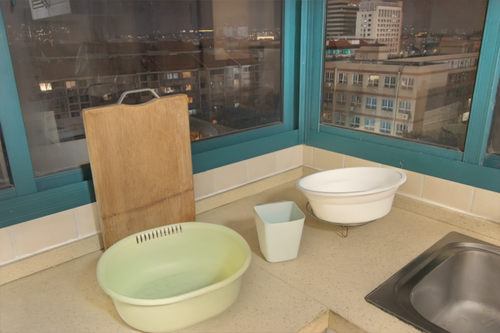} & 
    \includegraphics[clip=false, trim={0 0 0 0},width=0.19\columnwidth]{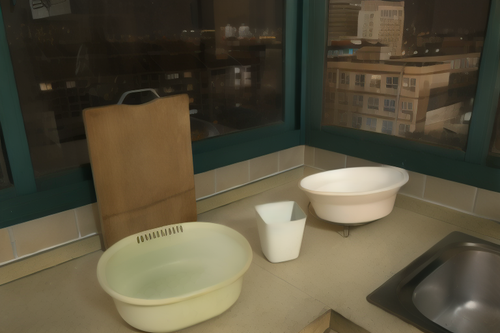} \\

    \midrule

        \includegraphics[clip=false, trim={0 0 0 0},width=0.19\columnwidth]{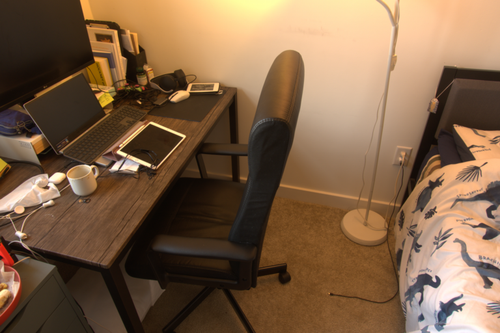} & 
    \includegraphics[clip=false, trim={0 0 0 0},width=0.19\columnwidth]{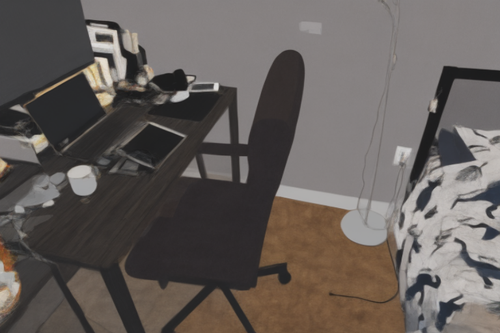} & 
    \includegraphics[clip=false, trim={0 0 0 0},width=0.19\columnwidth]{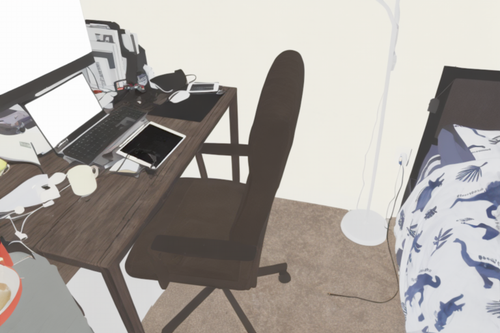} & 
    \includegraphics[clip=false, trim={0 0 0 0},width=0.19\columnwidth]{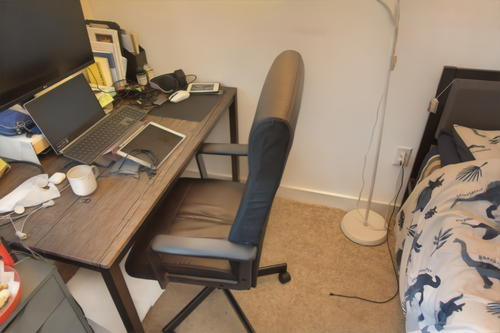} & 
    \includegraphics[clip=false, trim={0 0 0 0},width=0.19\columnwidth]{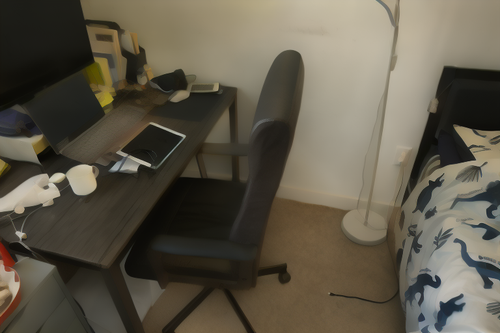} \\

    \includegraphics[clip=false, trim={0 0 0 0},width=0.19\columnwidth]{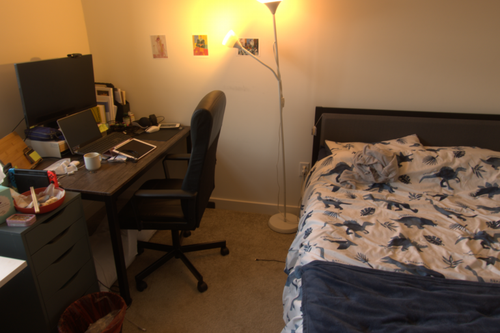} & 
    \includegraphics[clip=false, trim={0 0 0 0},width=0.19\columnwidth]{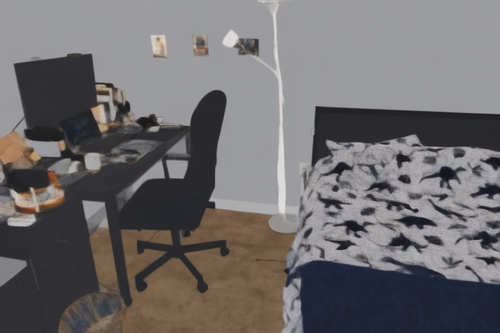} & 
    \includegraphics[clip=false, trim={0 0 0 0},width=0.19\columnwidth]{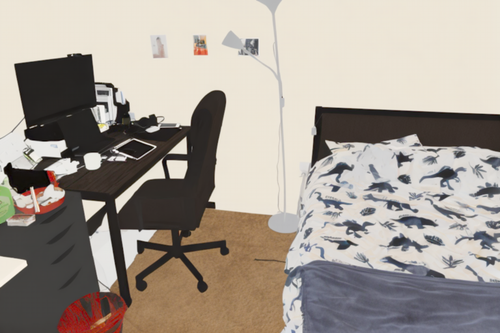} & 
    \includegraphics[clip=false, trim={0 0 0 0},width=0.19\columnwidth]{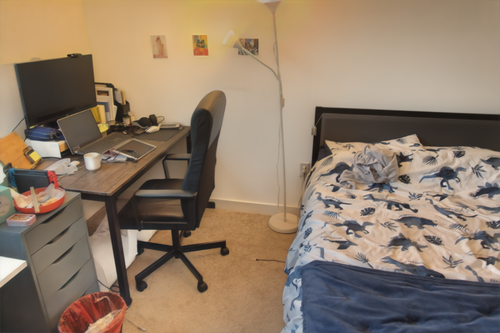} & 
    \includegraphics[clip=false, trim={0 0 0 0},width=0.19\columnwidth]{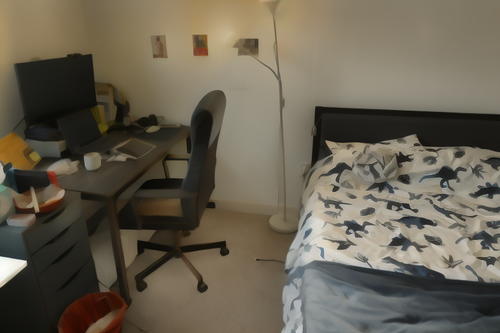} \\

\end{tabular}

\caption{We qualitatively compare the predicted albedos on the IIW dataset~\cite{bell2014intrinsic} and MAW dataset~\cite{wu2023measured}. Our method,~\method, produces albedos that preserve the true colors of objects and walls, without being biased by ambient light color. It also more effectively removes reflections and shadows, resulting in consistent albedo estimates under varying lighting conditions.}

\label{fig:iiw}

\end{figure*}

Figure~\ref{fig:iiw} shows a qualitative comparison of predicted albedos on the IIW and MAW datasets. Our method~\method~achieves more accurate color preservation, recovering the true colors of walls and objects without being biased by the ambient light color. While supervised methods~\cite{kocsis2023intrinsic, Zeng_2024} effectively remove lighting effects, they often distort object colors and tend to erase entire regions behind transparent or reflective surfaces like mirrors and windows. In contrast, the self-supervised baseline~\cite{zhang2024latent} struggles to remove ambient illumination, producing albedos with a strong color cast from the scene lighting. Our method overcomes these issues by producing more neutral and consistent albedos, even in scenes with complex lighting and reflections.\\

\begin{figure*}[tb] 

\centering

\def\fgsize{0.48}
\def\rowspacing{0.2cm}

\scriptsize
\setlength{\tabcolsep}{0.004\linewidth}
\renewcommand{\arraystretch}{1.0}
\begin{tabular}{ccccc}%

      Input  & IntrinsicDiffusion~\cite{kocsis2023intrinsic}  &$\text{RGB}  \to\text{X}$~\cite{Zeng_2024}&  LatentIntrinsics~\cite{zhang2024latent} &\method~(ours)\\

 \vspace{\rowspacing}

        \includegraphics[clip=false, trim={0 0 0 0},width=0.19\columnwidth]{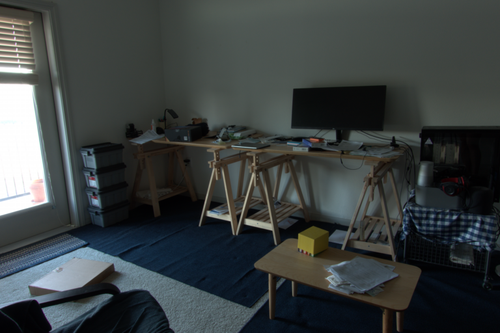} & 
    \includegraphics[clip=false, trim={0 0 0 0},width=0.19\columnwidth]{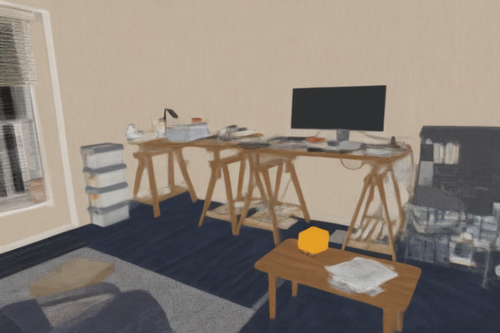} & 
    \includegraphics[clip=false, trim={0 0 0 0},width=0.19\columnwidth]{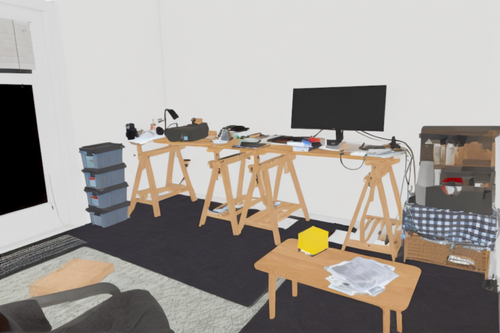} & 
    \includegraphics[clip=false, trim={0 0 0 0},width=0.19\columnwidth]{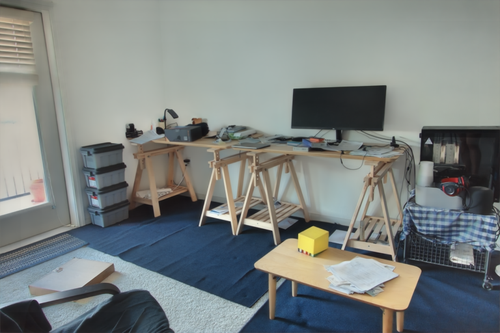} & 
    \includegraphics[clip=false, trim={0 0 0 0},width=0.19\columnwidth]{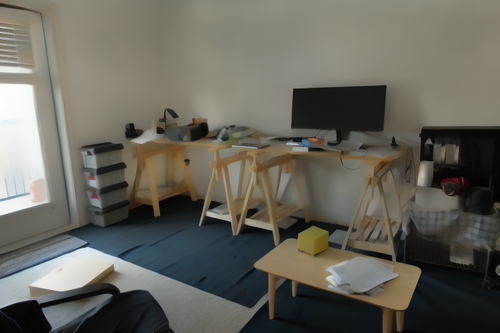} \\

        \includegraphics[clip=false, trim={0 0 0 0},width=0.19\columnwidth]{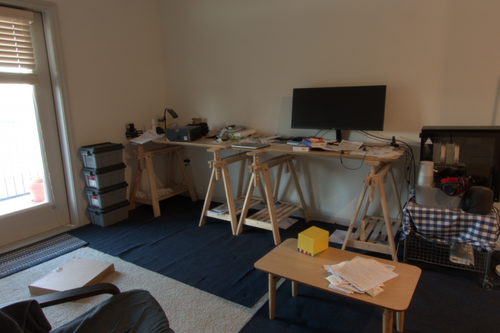} & 
    \includegraphics[clip=false, trim={0 0 0 0},width=0.19\columnwidth]{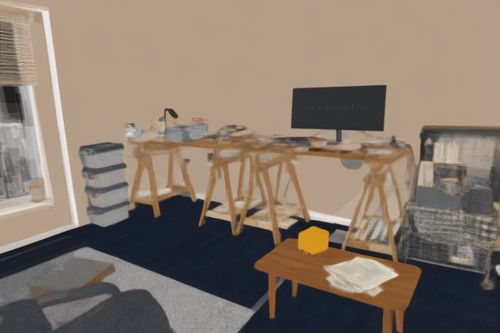} & 
    \includegraphics[clip=false, trim={0 0 0 0},width=0.19\columnwidth]{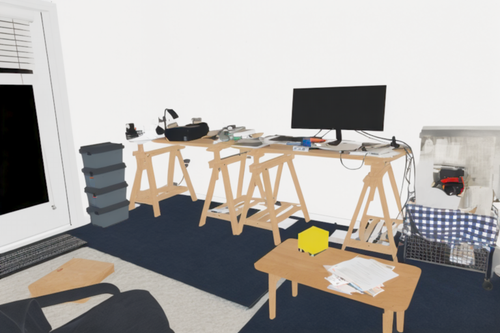} & 
    \includegraphics[clip=false, trim={0 0 0 0},width=0.19\columnwidth]{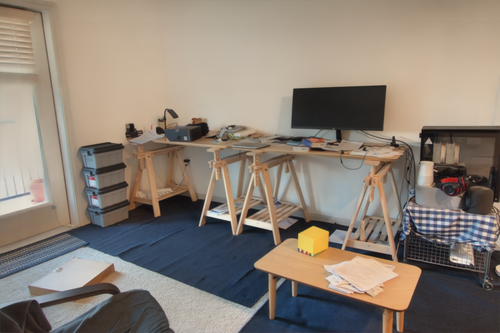} & 
    \includegraphics[clip=false, trim={0 0 0 0},width=0.19\columnwidth]{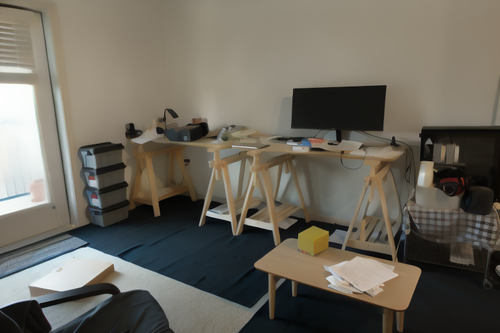} \\

        \includegraphics[clip=false, trim={0 0 0 0},width=0.19\columnwidth]{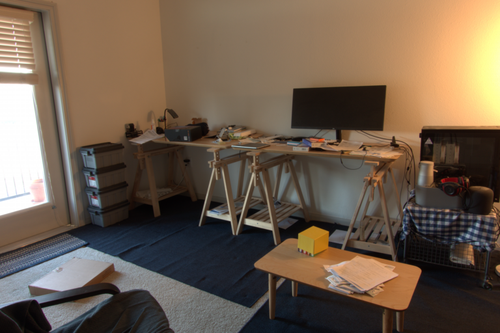} & 
    \includegraphics[clip=false, trim={0 0 0 0},width=0.19\columnwidth]{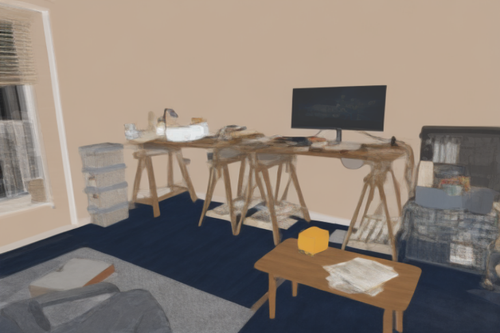} & 
    \includegraphics[clip=false, trim={0 0 0 0},width=0.19\columnwidth]{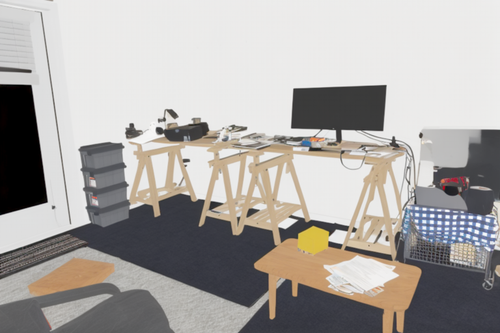} & 
    \includegraphics[clip=false, trim={0 0 0 0},width=0.19\columnwidth]{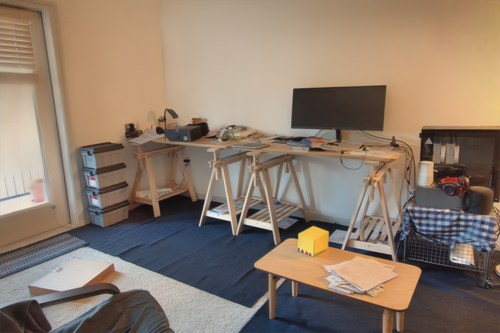} & 
    \includegraphics[clip=false, trim={0 0 0 0},width=0.19\columnwidth]{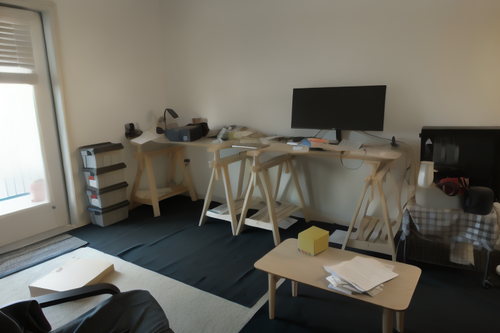} \\
            \includegraphics[clip=false, trim={0 0 0 0},width=0.19\columnwidth]{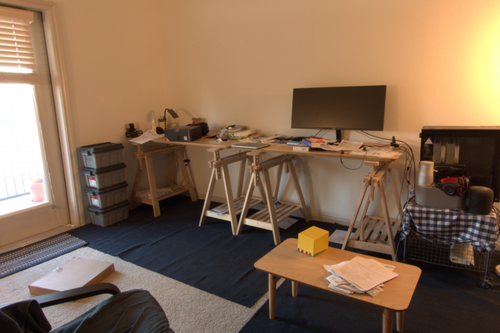} & 
    \includegraphics[clip=false, trim={0 0 0 0},width=0.19\columnwidth]{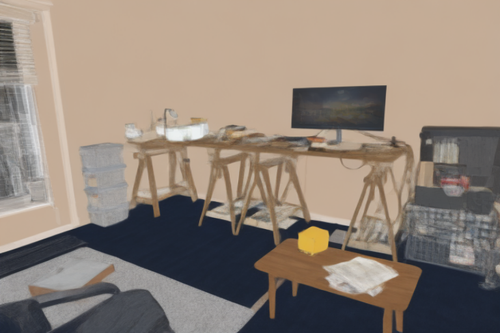} & 
    \includegraphics[clip=false, trim={0 0 0 0},width=0.19\columnwidth]{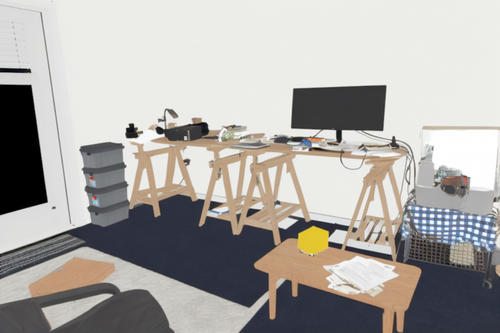} & 
    \includegraphics[clip=false, trim={0 0 0 0},width=0.19\columnwidth]{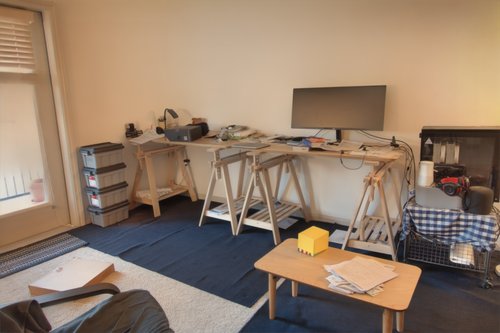} & 
    \includegraphics[clip=false, trim={0 0 0 0},width=0.19\columnwidth]{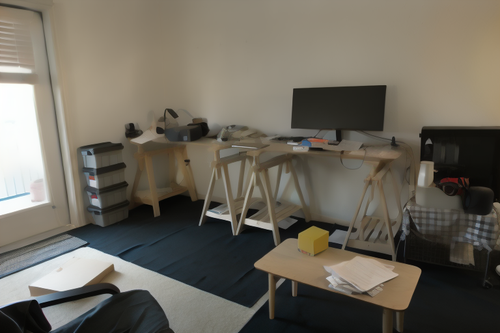} \\

    \includegraphics[clip=false, trim={0 0 0 0},width=0.19\columnwidth]{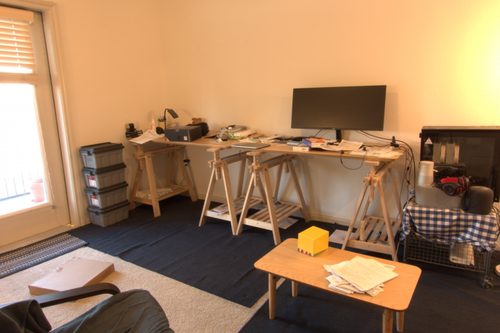} & 
    \includegraphics[clip=false, trim={0 0 0 0},width=0.19\columnwidth]{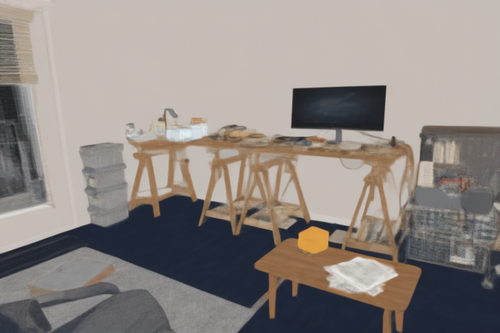} & 
    \includegraphics[clip=false, trim={0 0 0 0},width=0.19\columnwidth]{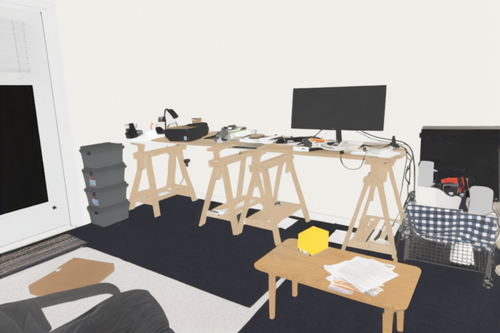} & 
    \includegraphics[clip=false, trim={0 0 0 0},width=0.19\columnwidth]{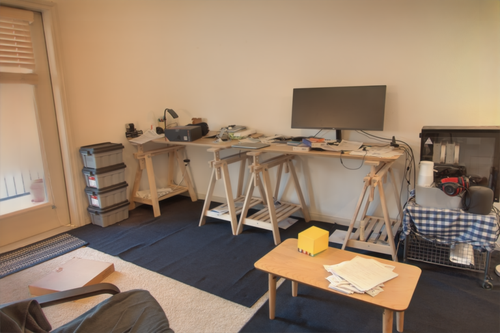} & 
    \includegraphics[clip=false, trim={0 0 0 0},width=0.19\columnwidth]{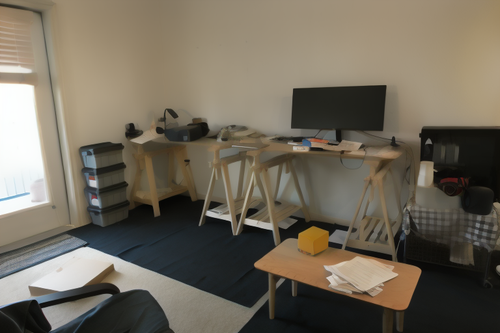} \\

\end{tabular}

  \caption{We qualitatively compare the predicted albedos on the MAW dataset~\cite{wu2023measured}. We show that~\method~predicts consistent albedos from the same scene under various lighting conditions.}
\label{fig:maw}

\end{figure*}

Figure~\ref{fig:maw} illustrates albedo predictions across multiple lighting conditions from a scene from the MAW dataset. Even subtle changes in light intensity lead to noticeable shifts in LatentsIntrinsics~\cite{zhang2024latent} output, which retains color biases from the scene illumination. IntrinsicDiffusion~\cite{kocsis2023intrinsic}, while better at handling lighting effects, still alters the wall colors across conditions. In contrast, \method~demonstrates strong consistency, producing stable albedos that remain close to the true surface colors regardless of the lighting variation. This highlights the robustness of our approach in disentangling intrinsic appearance from external illumination.

\begin{figure*}[tb] 
\centering
\def\fgsize{0.48}
\def\rowspacing{0.2cm}
\scriptsize
\setlength{\tabcolsep}{0.0035\linewidth}
\renewcommand{\arraystretch}{1.0}
\begin{tabular}{ccccc}%
      Input & \method~(ours)  & Relighting 1 &  Relighting 2 &  Relighting 3  \\ 

 \vspace{\rowspacing}
    \includegraphics[clip=false, trim={0 0 0 0},width=0.21\columnwidth]{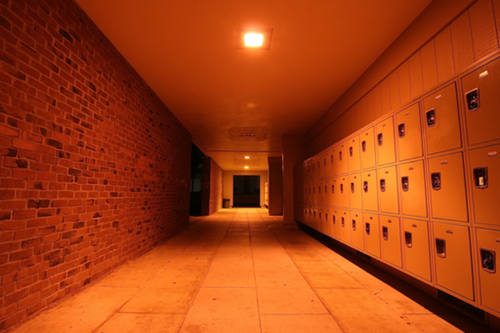} & 
        \includegraphics[clip=false, trim={0 0 0 0},width=0.21\columnwidth]{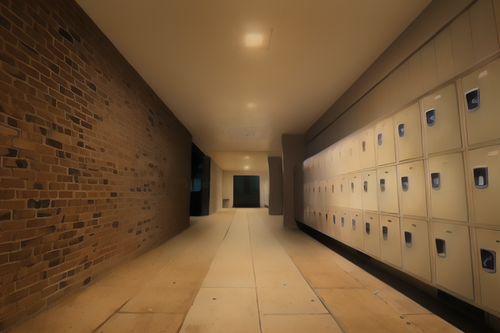} & 
    \includegraphics[clip=false, trim={0 0 0 0},,width=0.21\columnwidth]{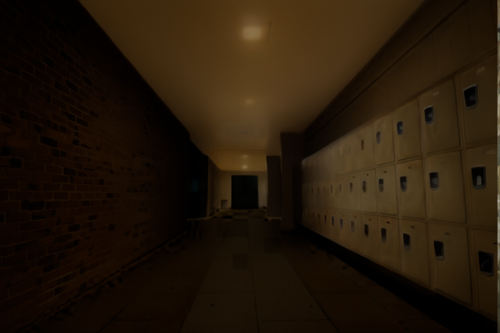} & 
    \includegraphics[clip=false, trim={0 0 0 0},,width=0.21\columnwidth]{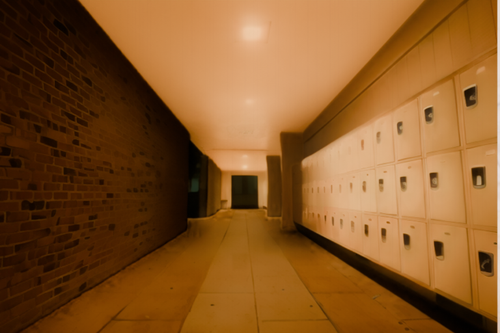} & 
    \includegraphics[clip=false, trim={0 0 0 0},width=0.21\columnwidth]{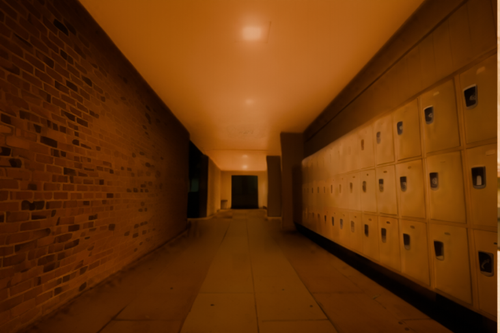} \\ 
        \includegraphics[clip=false, trim={0 0 0 0},width=0.21\columnwidth]{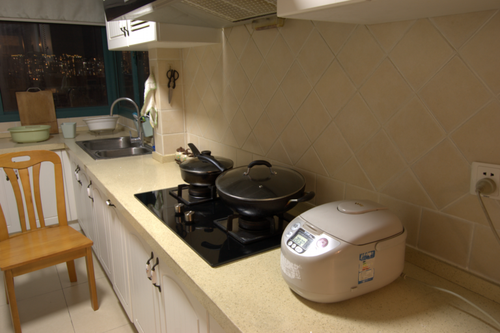} & 
        \includegraphics[clip=false, trim={0 0 0 0},width=0.21\columnwidth]{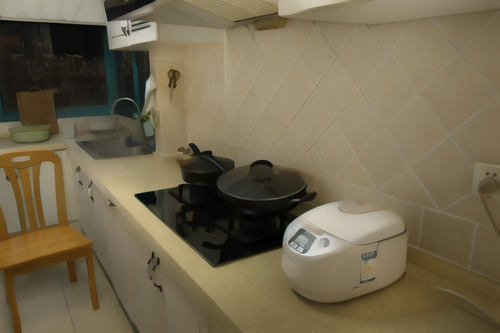} & 
    \includegraphics[clip=false, trim={0 0 0 0},,width=0.21\columnwidth]{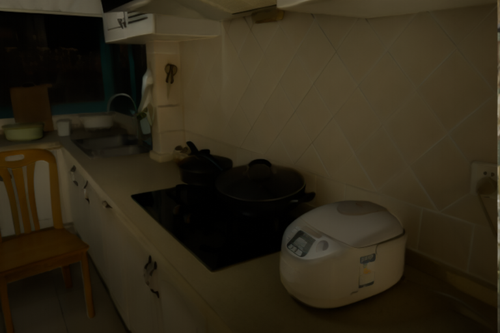} & 
    \includegraphics[clip=false, trim={0 0 0 0},,width=0.21\columnwidth]{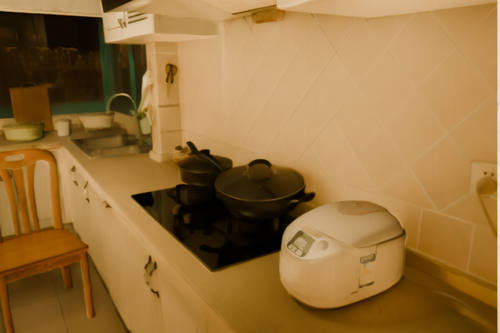} & 
    \includegraphics[clip=false, trim={0 0 0 0},width=0.21\columnwidth]{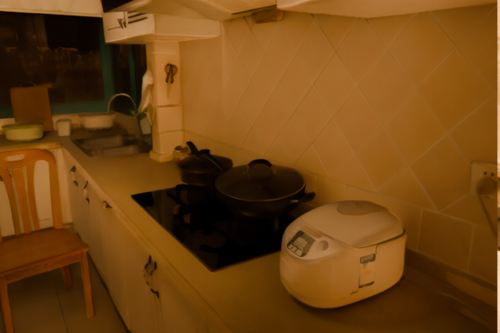} \\ 

\end{tabular}
\caption{We show results of virtual relighting with Blender~\cite{blender}, where we relight the predicted albedos with different environment maps.}
\label{fig:relighting}

\end{figure*}

\begin{figure*}[tb] 
\centering
\def\fgsize{0.48}
\def\rowspacing{0.2cm}
\scriptsize
\setlength{\tabcolsep}{0.0035\linewidth}
\renewcommand{\arraystretch}{1.0}
\begin{tabular}{ccccc}%
      Input & Predicted albedo  & Relighting 1 &  Relighting 2 &  Relighting 3  \\ 

 \vspace{\rowspacing}
    \includegraphics[clip=false, trim={0 0 0 0},width=0.21\columnwidth]{supps/input/10483.png} & 
        \includegraphics[clip=false, trim={0 0 0 0},width=0.21\columnwidth]{supps/ours/albedo_mean_10483.png} & 
    \includegraphics[clip=false, trim={0 0 0 0},,width=0.21\columnwidth]{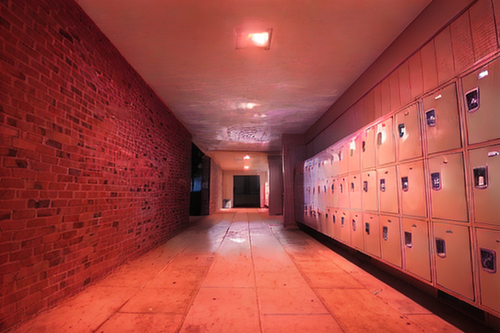} & 
    \includegraphics[clip=false, trim={0 0 0 0},,width=0.21\columnwidth]{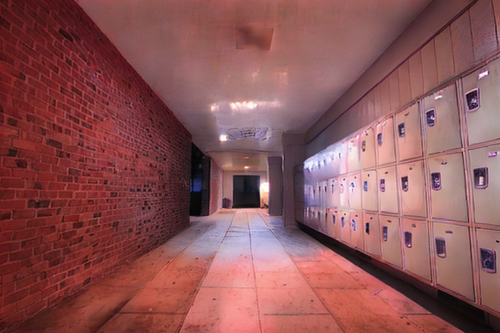} & 
    \includegraphics[clip=false, trim={0 0 0 0},width=0.21\columnwidth]{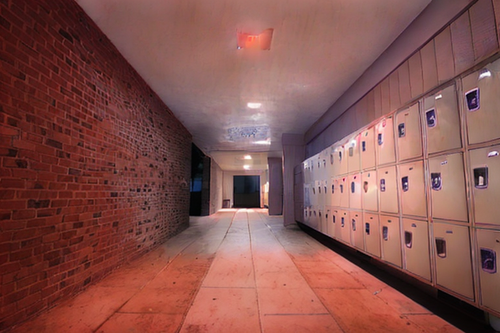} \\ 
        \includegraphics[clip=false, trim={0 0 0 0},width=0.21\columnwidth]{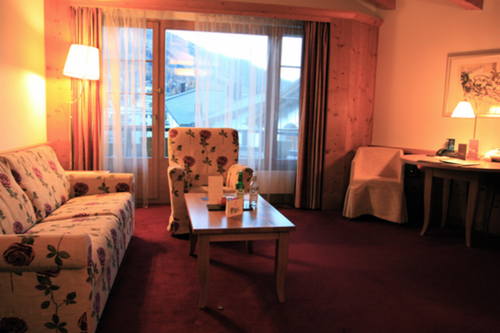} & 
        \includegraphics[clip=false, trim={0 0 0 0},width=0.21\columnwidth]{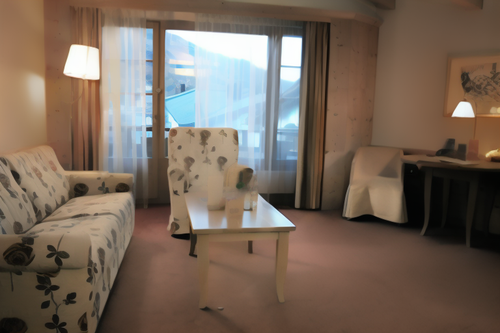} & 
    \includegraphics[clip=false, trim={0 0 0 0},,width=0.21\columnwidth]{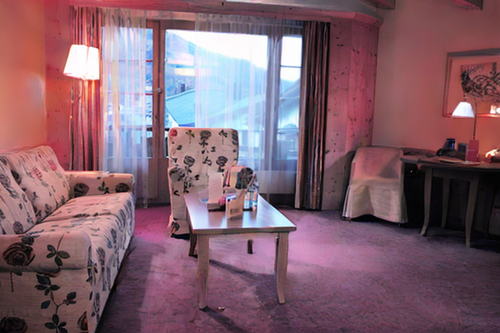} & 
    \includegraphics[clip=false, trim={0 0 0 0},,width=0.21\columnwidth]{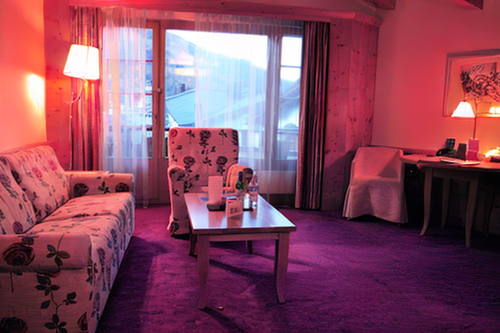} & 
    \includegraphics[clip=false, trim={0 0 0 0},width=0.21\columnwidth]{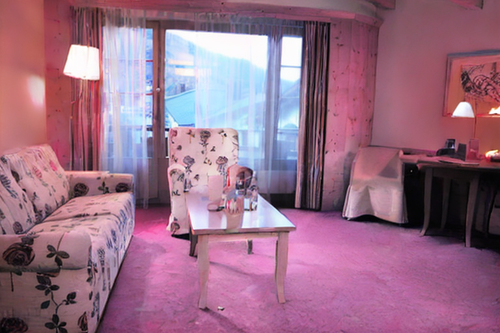} \\ 

            \includegraphics[clip=false, trim={0 0 0 0},width=0.21\columnwidth]{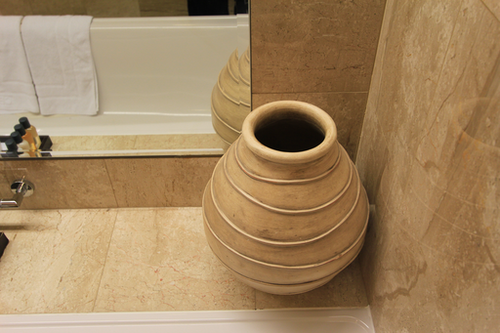} & 
        \includegraphics[clip=false, trim={0 0 0 0},width=0.21\columnwidth]{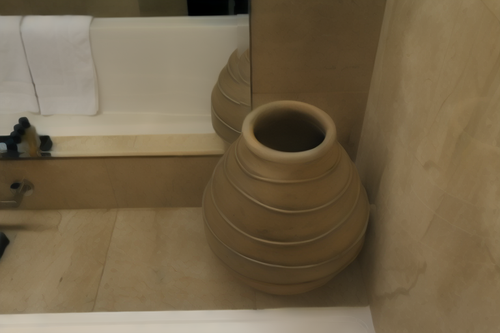} & 
    \includegraphics[clip=false, trim={0 0 0 0},,width=0.21\columnwidth]{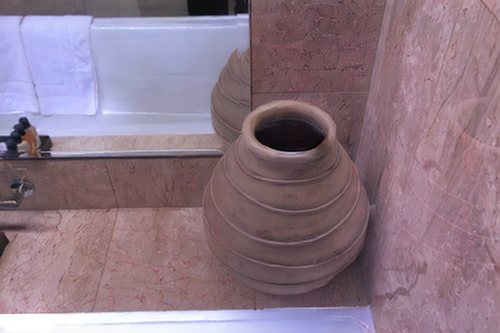} & 
    \includegraphics[clip=false, trim={0 0 0 0},,width=0.21\columnwidth]{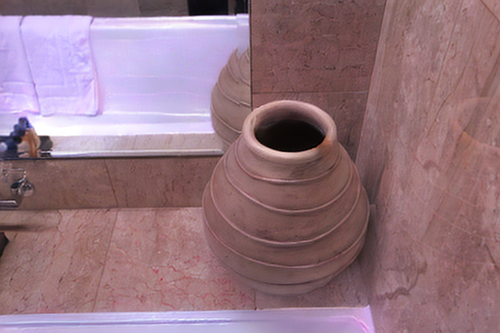} & 
    \includegraphics[clip=false, trim={0 0 0 0},width=0.21\columnwidth]{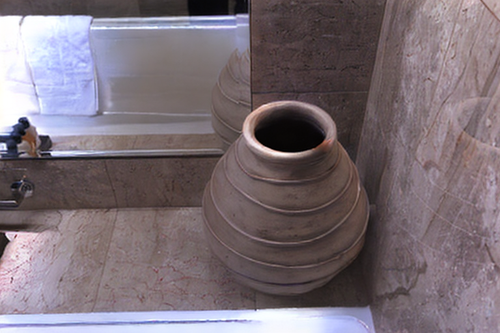} \\ 

\end{tabular}
\caption{We show unconditioned relighting results predicted by~\method~at different inferences. Each output corresponds to a different latent relighting sample $\hat{z}_i$, demonstrating the ability of \method~to generate diverse and plausible illumination conditions without any explicit supervision and conditioning}

\label{fig:ours-relighting}
\vspace{-5mm}
\end{figure*}

\section{Applications}

\paragraph{Unconditioned scene relighting}

We show in Fig.~\ref{fig:ours-relighting} unconditioned scene relighting as an application of \method's~predicted albedos. Our model predicts albedos that are not affected by illumination or ambient light color, allowing for an effective separation between light dependent and independent properties.\\
We sample different relighting latents $\hat{z}_i$ to generate multiple relit versions of the same input image. These relightings vary in ambient tone and illumination direction, producing realistic outputs under new lighting conditions. This illustrates the ability of our method to generalize relighting from a single image without any explicit supervision and conditioning.

\paragraph{Virtual scene relighting}
We demonstrate a downstream application of \method~ by relighting the predicted albedos in Blender\cite{blender} using environment maps downloaded from PolyHaven~\footnote{\url{https://polyhaven.com/}}. This experiment highlights the practical utility of our lighting-invariant albedo predictions. As shown in Fig.~\ref{fig:relighting}, our method produces albedos that preserve object appearance and enable consistent relighting results.

\clearpage
\bibliographystyle{plain}
\bibliography{main}

\begin{thebibliography}{10}

\bibitem{barrow1978recovering}
Harry Barrow, J~Tenenbaum, A~Hanson, and E~Riseman.
\newblock Recovering intrinsic scene characteristics.
\newblock {\em Comput. vis. syst}, 2(3-26):2, 1978.

\bibitem{bell2014intrinsic}
Sean Bell, Kavita Bala, and Noah Snavely.
\newblock Intrinsic images in the wild.
\newblock {\em ACM Transactions on Graphics (TOG)}, 33(4):1--12, 2014.

\bibitem{bhattad2022stylitgan}
Anand Bhattad, James Soole, and D.A. Forsyth.
\newblock Stylitgan: Image-based relighting via latent control.
\newblock In {\em Proceedings of the IEEE/CVF Conference on Computer Vision and Pattern Recognition (CVPR)}, pages 4231--4240, June 2024.

\bibitem{careagaIntrinsic}
Chris Careaga and Ya\u{g}{\i}z Aksoy.
\newblock Intrinsic image decomposition via ordinal shading.
\newblock {\em ACM Trans. Graph.}, 2023.

\bibitem{choi2023mair}
JunYong Choi, SeokYeong Lee, Haesol Park, Seung-Won Jung, Ig-Jae Kim, and Junghyun Cho.
\newblock Mair: multi-view attention inverse rendering with 3d spatially-varying lighting estimation.
\newblock In {\em 2023 IEEE/CVF Conference on Computer Vision and Pattern Recognition (CVPR)}, pages 8392--8401, 2023.

\bibitem{blender}
Blender~Online Community.
\newblock {\em Blender - a 3D modelling and rendering package}.
\newblock Blender Foundation, 2018.

\bibitem{du2023generative}
Xiaodan Du, Nicholas Kolkin, Greg Shakhnarovich, and Anand Bhattad.
\newblock Generative models: What do they know? do they know things? let's find out!
\newblock {\em arXiv preprint arXiv:2311.17137}, 2023.

\bibitem{du2021unsupervised}
Yilun Du, Shuang Li, Yash Sharma, Josh Tenenbaum, and Igor Mordatch.
\newblock Unsupervised learning of compositional energy concepts.
\newblock {\em Advances in Neural Information Processing Systems}, 34:15608--15620, 2021.

\bibitem{forsyth2021intrinsic}
David Forsyth and Jason~J Rock.
\newblock Intrinsic image decomposition using paradigms.
\newblock {\em IEEE transactions on pattern analysis and machine intelligence}, 44(11):7624--7637, 2021.

\bibitem{garces2022survey}
Elena Garces, Carlos Rodriguez-Pardo, Dan Casas, and Jorge Lopez-Moreno.
\newblock A survey on intrinsic images: Delving deep into lambert and beyond.
\newblock {\em International Journal of Computer Vision}, 130(3):836--868, 2022.

\bibitem{grosse2009ground}
Roger Grosse, Micah~K Johnson, Edward~H Adelson, and William~T Freeman.
\newblock Ground truth dataset and baseline evaluations for intrinsic image algorithms.
\newblock In {\em 2009 IEEE 12th International Conference on Computer Vision}, pages 2335--2342, 2009.

\bibitem{ho2022classifierfreediffusionguidance}
Jonathan Ho and Tim Salimans.
\newblock Classifier-free diffusion guidance, 2022.

\bibitem{kocsis2023intrinsic}
Peter Kocsis, Vincent Sitzmann, and Matthias Nie{\ss}ner.
\newblock Intrinsic image diffusion for indoor single-view material estimation.
\newblock {\em Proceedings of the IEEE/CVF conference on Computer Vision and Pattern Recognition}, pages 5198--5208, 2024.

\bibitem{laffont2015intrinsic}
Pierre-Yves Laffont and Jean-Charles Bazin.
\newblock Intrinsic decomposition of image sequences from local temporal variations.
\newblock In {\em Proceedings of the IEEE international conference on computer vision}, pages 433--441, 2015.

\bibitem{land1971lightness}
Edwin~H Land and John~J McCann.
\newblock Lightness and retinex theory.
\newblock {\em Journal of the Optical society of America}, 61(1):1--11, 1971.

\bibitem{li2018learning}
Zhengqi Li and Noah Snavely.
\newblock Learning intrinsic image decomposition from watching the world.
\newblock In {\em Proceedings of the IEEE conference on computer vision and pattern recognition}, pages 9039--9048, 2018.

\bibitem{li2020inverse}
Zhengqin Li, Mohammad Shafiei, Ravi Ramamoorthi, Kalyan Sunkavalli, and Manmohan Chandraker.
\newblock Inverse rendering for complex indoor scenes: Shape, spatially-varying lighting and svbrdf from a single image.
\newblock In {\em Proceedings of the IEEE/CVF conference on computer vision and pattern recognition}, pages 2475--2484, 2020.

\bibitem{liu2024review}
Siyuan Liu, Xiaoyue Jiang, Letian Liu, Zhaoqiang Xia, Sihang Dang, and Xiaoyi Feng.
\newblock A review of intrinsic image decomposition.
\newblock In {\em 2024 3rd International Conference on Image Processing and Media Computing (ICIPMC)}, pages 254--261, 2024.

\bibitem{loscos:inria-00527577}
C{\'e}line Loscos, Marie-Claude Frasson, George Drettakis, Bruce Walter, Xavier Granier, and Pierre Poulin.
\newblock {Interactive Virtual Relighting and Remodeling of Real Scenes}.
\newblock In {\em {Rendering Techniques '99}}, pages 329--340, 1999.

\bibitem{ma2018single}
Wei-Chiu Ma, Hang Chu, Bolei Zhou, Raquel Urtasun, and Antonio Torralba.
\newblock Single image intrinsic decomposition without a single intrinsic image.
\newblock In {\em Proceedings of the European conference on computer vision (ECCV)}, pages 201--217, 2018.

\bibitem{murmann19}
Lukas Murmann, Michael Gharbi, Miika Aittala, and Fredo Durand.
\newblock A multi-illumination dataset of indoor object appearance.
\newblock In {\em 2019 IEEE International Conference on Computer Vision (ICCV)}, 2019.

\bibitem{rombach2022high}
Robin Rombach, Andreas Blattmann, Dominik Lorenz, Patrick Esser, and Bj{\"o}rn Ommer.
\newblock High-resolution image synthesis with latent diffusion models.
\newblock In {\em Proceedings of the IEEE/CVF conference on computer vision and pattern recognition}, pages 10684--10695, 2022.

\bibitem{song2020denoising}
Jiaming Song, Chenlin Meng, and Stefano Ermon.
\newblock Denoising diffusion implicit models.
\newblock In {\em International Conference on Learning Representations (ICLR)}, 2021.

\bibitem{su2024compositionalimagedecompositiondiffusion}
Jocelin Su, Nan Liu, Yanbo Wang, Joshua~B. Tenenbaum, and Yilun Du.
\newblock Compositional image decomposition with diffusion models.
\newblock In {\em Proceedings of the 41st International Conference on Machine Learning}, pages 46823--46842, 2024.

\bibitem{weiss2001deriving}
Yair Weiss.
\newblock Deriving intrinsic images from image sequences.
\newblock In {\em Proceedings Eighth IEEE International Conference on Computer Vision. ICCV 2001}, volume~2, pages 68--75, 2001.

\bibitem{wu2023measured}
Jiaye Wu, Sanjoy Chowdhury, Hariharmano Shanmugaraja, David Jacobs, and Soumyadip Sengupta.
\newblock Measured albedo in the wild: Filling the gap in intrinsics evaluation.
\newblock In {\em 2023 IEEE International Conference on Computational Photography (ICCP)}, pages 1--12, 2023.

\bibitem{Zeng_2024}
Zheng Zeng, Valentin Deschaintre, Iliyan Georgiev, Yannick Hold-Geoffroy, Yiwei Hu, Fujun Luan, Ling-Qi Yan, and Miloš Hašan.
\newblock Rgb↔x: Image decomposition and synthesis using material- and lighting-aware diffusion models.
\newblock In {\em Special Interest Group on Computer Graphics and Interactive Techniques Conference Conference Papers ’24}, SIGGRAPH ’24, page 1–11. ACM, 2024.

\bibitem{zhang2024latent}
Xiao Zhang, William Gao, Seemandhar Jain, Michael Maire, David Forsyth, and Anand Bhattad.
\newblock Latent intrinsics emerge from training to relight.
\newblock {\em Advances in Neural Information Processing Systems}, 37:96775--96796, 2024.

\bibitem{zhu2023i2}
Jingsen Zhu, Yuchi Huo, Qi~Ye, Fujun Luan, Jifan Li, Dianbing Xi, Lisha Wang, Rui Tang, Wei Hua, Hujun Bao, et~al.
\newblock I2-sdf: Intrinsic indoor scene reconstruction and editing via raytracing in neural sdfs.
\newblock In {\em Proceedings of the IEEE/CVF conference on computer vision and pattern recognition}, pages 12489--12498, 2023.

\bibitem{zhu2022learning}
Jingsen Zhu, Fujun Luan, Yuchi Huo, Zihao Lin, Zhihua Zhong, Dianbing Xi, Rui Wang, Hujun Bao, Jiaxiang Zheng, and Rui Tang.
\newblock Learning-based inverse rendering of complex indoor scenes with differentiable monte carlo raytracing.
\newblock In {\em SIGGRAPH Asia 2022 Conference Papers}, pages 1--8, 2022.

\end{thebibliography}
\newpage

\end{document}